\documentclass[pdflatex,sn-mathphys-ay]{sn-jnl}

\usepackage{graphicx}%
\usepackage{multirow}%
\usepackage{amsmath,amssymb,amsfonts}%
\usepackage{amsthm}%
\usepackage[title]{appendix}%
\usepackage{xcolor}%
\usepackage{textcomp}%
\usepackage{comment}
\usepackage{xcolor}
\usepackage{booktabs}%
\usepackage{algorithm}%
\usepackage{algorithmicx}%
\usepackage{algpseudocode}%
\usepackage{listings}%
\usepackage{enumitem}
\usepackage{tabularx}%
\usepackage[normalem]{ulem}%
\DeclareRobustCommand{\add}[1]{{#1}}

\newcommand{\repourl}{\url{https://github.com/ProSoDiAC/agentic-ai-validation-mapping-study}}
\IfFileExists{xurl.sty}{\usepackage{xurl}}{}
\providecommand{\Description}[1]{}
\providecommand{\footnoteA}[1]{\footnote{#1}}
\AtBeginDocument{\hypersetup{hypertexnames=false}}

\makeatletter
\@ifundefined{@makespecialcolbox}{%
  \gdef\@makespecialcolbox{%
    \setbox\@outputbox \vbox{%
      \@texttop
      \dimen@ \dp\@outputbox
      \unvbox\@outputbox
      \vskip-\dimen@
    }%
    \@tempdima \@colht
    \ifdim \wd\@kludgeins>\z@
      \advance \@tempdima -\ht\@outputbox
      \advance \@tempdima \pageshrink
      \setbox\@outputbox \vbox to \@colht{%
        \unvbox\@outputbox
        \vskip \@tempdima
        \@textbottom
      }%
    \else
      \advance \@tempdima -\ht\@kludgeins
      \setbox \@outputbox \vbox to \@colht{%
        \vbox to \@tempdima{%
          \unvbox\@outputbox
          \@textbottom}%
        \vss}%
    \fi
    {\setbox \@tempboxa \box \@kludgeins}%
  }%
}{}
\@ifundefined{@make@specialcolbox}{\let\@make@specialcolbox\@makespecialcolbox}{}
\makeatother

\theoremstyle{thmstyleone}%
\theoremstyle{thmstyletwo}%

\theoremstyle{thmstylethree}%

\begin{document}

\title[Beyond Component Testing: Validating Agentic AI Systems]{Beyond Component Testing: Validating Agentic AI Systems}


\author[1,2]{\fnm{Fabio Orazio} \sur{Mirto}}\email{fmirto@unime.it}

\author[2]{\fnm{Luca} \sur{D'Agati}}\email{ldagati@unime.it}

\author[2]{\fnm{Giuseppe} \sur{Tricomi}}\email{gtricomi@unime.it}

\author*[3]{\fnm{Stefano} \sur{Silvestri}}\email{stefano.silvestri@icar.cnr.it}
\author[2,4]{\fnm{Francesco} \sur{Longo}}\email{flongo@unime.it}
\author[2,4]{\fnm{Antonio} \sur{Puliafito}}\email{apuliafito@unime.it}
\author[2,4]{\fnm{Giovanni} \sur{Merlino}}\email{gmerlino@unime.it}

\affil*[1]{\orgdiv{Department of Biomedical, Dental, and Morphological and Functional Imaging Sciences}, \orgname{University of Messina}, \orgaddress{\street{A.O.U. Policlinico “G.Martino” - Via Consolare Valeria}, \city{Messina}, \postcode{98125}, \country{Italy}}}

\affil[2]{\orgdiv{Department of Engineering}, \orgname{University of Messina}, \orgaddress{\street{Contrada di Dio,  Sant'Agata,  } \city{Messina}, \postcode{98158}, \country{Italy}}}

\affil[3]{\orgdiv{Institute for High Performance Computing and Networking of National Research Council of Italy }, \orgname{ICAR-CNR}, \orgaddress{\street{Via Pietro Castellino 111}, \city{Naples}, \postcode{80131},  \country{Italy}}}

\affil[4]{\orgdiv{National Interuniversity Consortium for Informatics}, \orgname{CINI}, \orgaddress{\street{ Via Ariosto, 25}, \city{Rome}, \postcode{00185},  \country{Italy}}}


\abstract{
Agentic AI systems act through multi-step trajectories that combine planning, tool use, memory, interaction, and adaptation. This behavior stretches validation practice beyond component testing and one-shot input--output evaluation, because acceptable system behavior now depends on how decisions unfold over time and under changing environmental conditions. \add{This systematic mapping study} synthesizes \add{262} papers spanning agent evaluation, software assurance, cyber-physical systems, runtime monitoring, and regulatory guidance in order to characterize the validation problem for agentic systems. The review is organized around a five-dimension taxonomy covering behavioral, safety, temporal, regulatory, and multi-agent concerns, and uses that taxonomy to map current approaches and \add{identify dense and sparse pairings of approaches and dimensions}. \add{The resulting map shows that the literature concentrates on behavioral evaluation (105 of 262 papers) and is thinnest on temporal validity (14 papers); regulatory work rests mainly on assurance cases and regulatory analysis, largely from IEEE-indexed venues, and multi-agent work mainly on benchmarks. }
Three cross-domain case studies (medical care, industrial operations, smart-mobility systems) provide operational illustrations of how the five taxonomy dimensions recur in safety-critical settings, motivated by the failure patterns documented in the reviewed literature. The paper concludes with a lifecycle-oriented research agenda centered on bounded-autonomy specifications, adversarial trajectory generation, runtime monitoring, and audit-ready evidence structures. The central claim is that trustworthy deployment of agentic AI depends on validating trajectories in context rather than assessing isolated components alone.
}




\maketitle
\section{Introduction}

\subsection{The Agentic Shift}

Agentic artificial intelligence (AI) systems are increasingly deployed as software systems that pursue goals through multi-step reasoning, planning, memory, tool use, and adaptation under changing context~\citep{luo2025,zou2025}. This shift changes the validation target. Classical software components primarily implement bounded functions over inputs, state, and interfaces, whereas agentic systems must be evaluated as policies acting over trajectories in dynamic environments. The question is therefore no longer only whether a component returns the correct output, but whether the overall system behaves acceptably over time, across interactions, and under changing operational conditions.

Recent agent architectures make this shift concrete. Contemporary frameworks organize execution around loops of perception, planning, tool invocation, reflection, memory update, and possible delegation to other agents or services~\citep{autogen2025,luo2025}. Even when built from familiar software parts, such as models, Application Programming Interfaces (APIs), retrieval modules, schedulers, and user interfaces, their system-level behavior emerges from how these parts are orchestrated across trajectories rather than from any single module in isolation. The result is a software architecture that is more open-ended, history-dependent, and environment-coupled than the systems for which mainstream testing abstractions were originally designed.

\subsection{Five Testing Mismatches}

Software engineering already provides mature foundations for unit and regression testing~\citep{myers2004,beizer1990}, as well as integration testing and specification-based verification~\citep{ammann2016}. Those foundations remain necessary. However, they were largely developed for systems whose relevant behaviors are relatively reproducible, decomposable, and specifiable at the level of components and interfaces. Agentic systems stress those assumptions in at least five recurring ways.

\begin{table*}[!tbp]
\centering
\caption{Five abstract mismatches between classical testing assumptions and agentic-system validation targets.}
\label{tab:intro-mismatch}
\footnotesize
\setlength{\tabcolsep}{4pt}
\renewcommand{\arraystretch}{1.05}
\begin{tabularx}{\textwidth}{@{}>{\raggedright\arraybackslash}p{2.3cm}>{\raggedright\arraybackslash}X>{\raggedright\arraybackslash}X>{\raggedright\arraybackslash}X@{}}
\toprule
\textbf{Mismatch} & \textbf{Classical testing assumption} & \textbf{Agentic-system reality} & \textbf{Validation implication} \\\midrule
Determinacy mismatch & Repeating the same test should yield the same or predictably equivalent result. & Stochastic inference, contingent planning, and tool availability can produce different trajectories from the same starting state. & Validation must characterize acceptable variation, not only exact reproducibility. \\\midrule
Decomposition mismatch & \add{Component-level evidence, complemented by integration tests over a bounded set of interactions, supports system-level claims.} & Failures often arise from orchestration across planner, memory, tools, users, and other agents. & Evidence must cover end-to-end trajectories and cross-component interactions. \\\midrule
Specification mismatch & Requirements can be stated as explicit input–output properties or interface contracts. & Goals such as ``act helpfully,'' ``escalate appropriately,'' or ``use tools safely'' are partial, contextual, and temporally extended. & Validation needs bounded-autonomy contracts, runtime constraints, and scenario-based evidence. \\\midrule
Environment mismatch & Operating conditions can be bounded into stable test fixtures and controlled mocks. & Agent behavior depends on dynamic environments, external services, changing data, and human responses. & Assurance must include simulation, adversarial scenarios, and monitoring in context. \\\midrule
Temporal mismatch & Correctness is treated as a property established prior to deployment. & Agentic behavior unfolds over long horizons and can drift with memory, updates, and changing workflows. & Validation must become lifecycle-oriented, with post-deployment oversight and evidence refresh. \\\bottomrule
\end{tabularx}
\end{table*}

\add{The second column states the assumptions of simplified, component-focused testing; Section~\ref{sec:core-assumptions} discusses how classical assurance practice qualifies them. }Table~\ref{tab:intro-mismatch} shows that unit, regression, integration, and specification-based tests cover only part of the assurance space. The validating question expands from ``Did the component return the correct output?'' to ``What evidence justifies trust in this system's behavior over realistic trajectories and operating conditions?'' Established software-engineering practice therefore remains essential but requires extension for agentic autonomy.

Section~\ref{sec:classical} returns to these same five abstract mismatches by examining them across classical testing methods. In that comparison, fault injection is listed separately because it operationalizes both decomposition and environment concerns within the five mismatches of Table~\ref{tab:intro-mismatch}.

\subsection{Why a Validation-Centered Survey Now}

Three developments make a validation-centered synthesis timely. The first is a gap in the survey landscape. The agent surveys of 2024--2026 map architectures, capabilities, benchmarks, and evaluation practice with increasing precision~\citep{yehudai2025,luo2025}, but they organize the field by what agents \emph{can do}, not by what must be \emph{shown} before and during consequential deployment. As Table~\ref{tab:survey-comparison} makes explicit, temporal validity and regulatory legibility are rarely treated as first-class axes among the closest adjacent surveys, and \add{only two rest} on a systematically screened corpus. A reader of the existing literature can learn how agents are built and how well they score; what evidence package would justify trusting one in deployment is a question that literature does not yet organize itself around.

The second development is that the governance clock is already running. \add{These sources differ in legal authority. In the European Union (EU), the Medical Devices Regulation (MDR) imposes binding post-market surveillance and documentation obligations~\citep{eu-mdr}; MDCG guidance, although not legally binding, explains how these obligations apply to all devices~\citep{mdcg2025-10} and how they interact with the AI Act for AI-based device software~\citep{mdcg2025-6}. In the US, FDA guidance on predetermined change control plans for AI-enabled devices is final~\citep{fda-pccp2025}, whereas its lifecycle-management guidance for AI-enabled device software functions remains a non-binding draft~\citep{fda-aiedsflm2025}. Taken together, they make lifecycle evidence, traceability, and change control explicit expectations for AI-enabled medical devices. These documents state what is expected;} the engineering literature that should supply the demonstration methods remains fragmented across the five streams surveyed here. The cost of that fragmentation is borne now, by teams deploying agentic systems in clinical, industrial, and mobility settings under evidence expectations that the technical literature does not yet operationalize.

The third development is the rapid consolidation of the empirical base: 209 of the \add{262} included papers \add{(79.8\%)} were published in 2025--2026 (Figure~\ref{fig:year-dist})\add{; Section~\ref{sec:methodology-limits} examines how this recency affects the stability of the conclusions}. This concentration makes it possible to synthesize agent-specific validation work while evaluation norms are still developing.

\subsection{Contributions}

\add{This study} makes four contributions:
\begin{enumerate}
    \item \textbf{\add{Systematic mapping study.}} We synthesize five partially disconnected bodies of literature: classical software engineering (SE) testing, agent-framework evaluation, Cyber-Physical Systems (CPS) validation, runtime assurance, and regulatory guidance, through a \add{documented }screening of \add{4,767 unique} records \add{(plus 49 candidates identified by other methods) }down to \add{262} included papers\add{; the study maps the distribution of coded publications and does not appraise study quality (Section~\ref{sec:review-design})}.

    \item \textbf{Five-dimension taxonomy.} We introduce a standalone taxonomy of validation challenges for agentic systems spanning behavioral, safety, temporal, regulatory, and multi-agent dimensions, with explicit validation objects, characteristic failure modes, and measurable metrics for each. Its distinguishing choice is to treat temporal validity and regulatory legibility as first-class axes rather than as remarks appended to capability valuation, the two axes on which adjacent surveys are weakest (Section~\ref{sec:taxonomy}).

    \item \textbf{\add{Quantified coverage map}.} We \add{map} the coded corpus: a coverage matrix over approach families and validation dimensions \add{shows which pairings are densely and which are thinly represented in the coded corpus}, with \add{sensitivity analyses over publication status, retrieval source, primary-dimension coding confidence, publication year, and rater disagreement that indicate which directional claims remain stable} (Section~\ref{sec:approaches}).

    \item \textbf{Research agenda.} We \add{develop, as an interpretive synthesis of the coverage map and of the surveyed literature,} a four-direction validation stack with candidate metrics and software-engineering targets, moving from a generic call for better evaluation toward a concrete research program (Section~\ref{sec:agenda}).
\end{enumerate}

The remainder of the paper is organized as follows. Section~\ref{sec:methodology} describes the survey methodology. Section~\ref{sec:related} organizes prior literature along five validation axes. Section~\ref{sec:definitions} defines agentic systems and the expanded assurance target, while Section~\ref{sec:classical} analyzes which classical software-testing assumptions break and which still carry over. Section~\ref{sec:taxonomy} presents the five-dimension taxonomy, and Section~\ref{sec:approaches} maps existing approaches onto it to identify directional gaps. Section~\ref{sec:casestudy} grounds the taxonomy in three consequential domains, Section~\ref{sec:agenda} develops the research agenda, Section~\ref{sec:challenges} discusses open challenges, and Section~\ref{sec:conclusion} concludes.

\subsection{Evidence Layers Used in This Paper}
\label{sec:evidence-layers}

\begingroup
The manuscript combines four kinds of statements whose evidential status differs, and we mark them consistently in the text, tables, and captions (Table~\ref{tab:evidence-layers}). Only the first layer is a finding that follows directly from the reviewed corpus; the other three are the authors' construction and weighed accordingly.

\begin{table}[!htbp]
\centering
\caption{Evidence layers in the manuscript and how each should be read.}
\label{tab:evidence-layers}
\footnotesize
\setlength{\tabcolsep}{3pt}
\renewcommand{\arraystretch}{1.15}
\begin{tabularx}{\columnwidth}{|>{\raggedright\arraybackslash}p{2.0cm}|>{\raggedright\arraybackslash}X|>{\raggedright\arraybackslash}p{1.6cm}|}
\hline
\textbf{Layer} & \textbf{Content and status} & \textbf{Where} \\
\hline
Corpus-supported findings & Counts, distributions, coverage matrix, and sensitivity analyses over the 262 coded papers. Descriptive and volume-based; conditional on single-reviewer screening and on the retrieval set. & Sec.~\ref{sec:methodology}, \ref{sec:approaches} \\
\hline
Conceptual taxonomy & Five validation dimensions and their sub-facets. Author synthesis; heuristic and overlapping; validated only for codability (inter-rater $\kappa$), not for completeness or independence. & Sec.~\ref{sec:taxonomy} \\
\hline
Normative recommendations & Research agenda, candidate metrics, artifacts. Interpretive; not derived by inference from the counts alone. & Sec.~\ref{sec:agenda}, \ref{sec:challenges} \\
\hline
Constructed case studies & Step-by-step traces combining literature-documented mechanisms with constructed defects and illustrative thresholds; not empirical evidence. & Sec.~\ref{sec:casestudy} \\
\hline
\end{tabularx}
\end{table}
\endgroup

\section{Survey Methodology}
\label{sec:methodology}

\subsection{Research Questions}

The review is structured by four research questions: which classical software-testing assumptions fail for agentic systems (RQ1), which validation dimensions a comprehensive framework must cover (RQ2), how existing approaches distribute across those dimensions (RQ3), and which research directions follow from the remaining gaps (RQ4). These RQs map directly onto Sections~\ref{sec:classical},~\ref{sec:taxonomy}, ~\ref{sec:approaches}, and~\ref{sec:agenda}.

\subsection{Review Design and Taxonomy Chronology}
\label{sec:review-design}

\begingroup
This study is a PRISMA-inspired \emph{systematic mapping study}~\citep{petersen2015mapping,kitchenham2007guidelines,page2021prisma}: search, eligibility, and coding procedures are explicit and reported, but there is no synthesis of effects and no study-quality assessment, which mapping-study guidelines do not require because the aim is to classify and count the literature rather than to aggregate its results~\citep{petersen2015mapping,kitchenham2007guidelines}; screening was performed by a single reviewer. Its corpus-level claims therefore concern the \emph{distribution of coded publications}, not the strength of the evidence those publications provide; terms such as ``gap'' and ``mature'' are used only in this volume-based sense.

The taxonomy was developed in a \emph{hybrid} (deductive-then-inductive) sequence. The five dimensions were first defined a priori from the assurance, cyber-physical, and regulatory literatures; these definitions informed dimension-specific search terms; a pilot coding round on a subset of records sharpened the operational definitions\add{, with no dimension added or removed afterwards; the definitions were then restated by validation object, the four borderline rules were written into the codebook (Section~\ref{sec:coding-rules} and the released codebook), and this version was frozen before full-corpus coding.} \label{q:chronology}In the frozen codebook each dimension is defined by its validation object, and all four borderline rules are part of it; both raters coded all 262 papers with this version, independently of each other, and the intra-rater audit (Section~\ref{sec:coding-scheme}) used the same version. The core query contains no dimension-specific vocabulary; such vocabulary appears only in the IEEE Xplore query (Section~\ref{sec:search-strategy}), which supplies half of the single-source records. The distributions reported below therefore describe coverage \emph{within the retrieved space}, and circularity between taxonomy and search is treated as a threat to validity (Section~\ref{sec:methodology-limits}).
\endgroup

\subsection{Search Strategy}
\label{sec:search-strategy}
\enlargethispage{\baselineskip}

We searched five primary databases: ACM Digital Library, IEEE Xplore, Scopus, arXiv, and Semantic Scholar. The initial search was conducted between January and April 2025, followed by a first refresh pass \add{between January and April 2026} that re-interrogated the same five sources to capture newly indexed 2025--2026 publications\add{ (the last Scopus search ran on 25 April, and the arXiv results extend to 15 April)}\add{, and by a second refresh pass in September 2026 that repeated the queries on the same five sources, with IEEE Xplore searched through the query described below}.\add{ After export, a date filter was applied to the records of all five sources, including those retrieved by both refresh passes: only records first made public between 1 January 2019 and 31 March 2026 were retained (criterion I5, Section~\ref{sec:inclusion-criteria}).} \add{\label{q:dates}Three dates are kept apart. \emph{Eligibility} depends only on the date of first public availability: the first online (early-access) date for journal papers, the earlier of the conference date and the first online date for conference papers, and the announcement date of the first arXiv version for preprints. The \emph{recorded version} of each study (DOI, venue, and year) is its version of record if that was published by 31 March 2026, the end of the inclusion window, and otherwise its arXiv version; a study first made public within the window whose version of record appeared later therefore remains eligible and is recorded as a preprint. \emph{Publication status} (published or preprint) refers to the same cut-off: a study counts as published only if its peer-reviewed version was available by 31 March 2026, and versions of record that appeared later are not used. The year bands reported below use the year of the recorded version.} The review covers publications from January 2019 to March 2026.\add{ Earlier foundational work is cited only as context, and no pre-2019 paper is in the corpus.} Unless noted otherwise, counts reported below refer to the consolidated retrieval set after \add{all} passes.

\add{The core query combines agent-identity terms (``agentic,'' ``LLM agent,'' ``language model agent,'' ``autonomous agent''), validation terms (``validation,'' ``assurance,'' ``verification,'' ``testing,'' ``evaluation,'' ``monitoring''), and trajectory-level terms (``trajectory,'' ``tool use,'' ``multi-step,'' ``memory,'' ``stateful''), each source using a subset of these terms in its own field syntax. This core query was run on the ACM Digital Library (abstract), Scopus (title, abstract, and keywords), Semantic Scholar, and arXiv (title and abstract field prefixes, entered in the arXiv simple search without a subject-category filter). \label{q:ieeequery}Two further queries use a broader formulation. The IEEE Xplore query replaces the trajectory terms with agent-identity terms in the title and a broad set of assurance-related terms in the abstract (safety, security, trust, governance, compliance, verification, benchmark, reliability, monitoring, accountability, robustness, drift, runtime), and adds clauses specific to each dimension: drift (temporal), assurance and safety cases and runtime assurance (regulatory and safety), multi-agent coordination and governance (multi-agent), and agentic benchmarks (behavioral). The IEEE Xplore records analyzed in this review come from the September 2026 export of this query, to which the date filter of criterion I5 was applied as for all sources. A supplementary arXiv query (``LLM agent'' evaluation) broadened the arXiv retrieval. Together, these two queries account for 3{,}559 of the 5{,}187 retrieved records. All query strings are released with the data (see Data availability).}

We additionally performed forward and backward citation searches from agent-evaluation anchor papers~\citep{luo2025,yehudai2025} and \add{healthcare-agent review anchor papers}~\citep{collaco2026,zhao2026}, and included regulatory documents from the Food and Drug Administration (FDA) and Medical Device Coordination Group (MDCG) not indexed in academic databases.

All retrieved records were exported with bibliographic metadata (including title, abstract, venue, year, and DOI where available) and consolidated into a unified dataset for screening and analysis.

\add{The five sources returned 5{,}187 records in total (arXiv 1{,}256; ACM Digital Library 808; IEEE Xplore 2{,}414; Scopus 306; Semantic Scholar 403). Removing 311 records outside the 2019--March 2026 window (among them 156 from Semantic Scholar, whose query had no date constraint, and 71 arXiv preprints whose first version was announced in April 2026) and 8 duplicates within the arXiv results left 4{,}868; title-based deduplication across sources removed a further 101, yielding \textbf{4{,}767 unique records}. Of these, 4{,}669 carry a single-source provenance tag (IEEE Xplore 2{,}369; arXiv 1{,}104; ACM Digital Library 708; Scopus 270; Semantic Scholar 218) and 98 were retrieved from two or more sources. Regulatory and standards documents from the FDA and MDCG were retained as background references but are not counted in the screened paper corpus. IEEE Xplore supplies 50.7\% of single-source records, about twice the arXiv share (23.6\%); Section~\ref{sec:methodology-limits} analyzes the effect of this skew.}

Deduplication was performed using title \add{}normalization, followed by manual inspection for ambiguous cases and cross-source matches.\add{ \label{q:vor}When a preprint and a published version of the same study were both retrieved, only one record was retained: the version of record if it was available by 31 March 2026, and otherwise the preprint (see the recorded-version rule above).}

\subsection{Inclusion and Exclusion Criteria}
\label{sec:inclusion-criteria}

\add{The following criteria governed both selection stages (Section~\ref{sec:paper-selection}).} \add{They were defined before screening began and applied to all records.}

\textbf{Inclusion criteria.} A paper was included if it satisfied all
of the following conditions:
\begin{itemize}
    \item \textbf{I1: Validation relevance.} The primary contribution had to address validation, testing, verification, runtime monitoring, assurance, benchmarking, safety enforcement, or auditable oversight of software behavior.
    \item \textbf{I2: Agentic execution feature.} The evaluated system had to execute over a trajectory rather than a single inference step. For corpus-inclusion purposes, this required \emph{at least one agentic execution feature}: explicit planning across steps, tool invocation with feedback, persistent memory/state carry-over, closed-loop interaction with an external environment, or coordination among multiple autonomous components. These disjunctive criteria ensured that papers addressing trajectory-level validation in any form could enter the corpus, even if they did not emphasize all three canonical agentic properties.
    \item \textbf{I3: Software-level scope.} The paper had to contain a substantive software or systems contribution relevant to agentic AI, CPS, runtime assurance, or AI-enabled software engineering.
    \item \textbf{I4: Sufficient technical content.} Empirical studies, benchmark papers, frameworks, formal methods papers, and technical
    reports were included only if they supplied enough methodological detail to support analytical coding.
    \item \textbf{I5: Time and language window.} The source had to be published in English and fall within the January~2019 to
    March~2026 search window.\add{ The date used is the date of first public availability: the first online (early-access) date for journal articles, not the date of the print issue; for conference papers, the earlier of the conference date and the first online date; and, for arXiv preprints, the announcement date of the first version, not the date of a later version (Section~\ref{sec:search-strategy}).}
\end{itemize}

\textbf{Exclusion criteria.} A paper was excluded if any of the
following conditions held:
\begin{itemize}
    \item \textbf{E1: Single-shot inference only.} The paper evaluated a single forward pass, a static prompt-response exchange, or prompt quality without an autonomous trajectory.
    \item \textbf{E2: Prompt-only orchestration.} The system used prompt templates, self-consistency, or prompt ensembles but did not expose persistent state, tool-mediated feedback, iterative control, or interacting agents.
    \item \textbf{E3: Non-agentic adjacent AI.} The contribution focused on perception-only natural language processing (NLP), vision, or prediction tasks not embedded in an autonomous action loop.
    \item \textbf{E4: Non-software assurance target.} The paper addressed hardware reliability, sensor fault tolerance, or physical device dependability without a software-level validation contribution.
    \item \textbf{E5: Insufficient scholarly record.} Editorials, blog posts, slide decks, and very short abstracts without retrievable technical content were excluded. \add{Preprints were assessed against the same criteria as peer-reviewed records; a preprint was excluded under E5 or I4 when its content was too preliminary to support coding.}
\end{itemize}

\begingroup
\label{q:criteria}In applying the criteria, I2 operated as an eligibility filter, not as a classification: a record qualified if the evaluated system exhibited at least one agentic execution feature \emph{and} the validation contribution addressed trajectory-level behavior. Whether a qualifying paper concerns a canonical agentic system, a boundary case, or a consequential assurance trigger (Section~\ref{sec:definitions}) was not recorded per paper, so corpus-level findings refer to the corpus as a whole. Three corpus papers illustrate the distinction: a multi-agent LLM system for industrial drive commissioning that uses expert tools, acts on an external environment, and keeps a human in the loop is a canonical case~\citep{driveaiagentMultiagent2025}; a runtime risk-assessment framework for autonomous cyber-physical systems, whose agentic feature is closed-loop interaction without LLM-style tool use or planning, is a boundary case~\citep{resonateRuntime2025}; and an interlocking safety-case approach for UAVs in shared airspace, evaluated in simulation and in an outdoor field study, is a consequential assurance trigger~\citep{interlockingSafety2025}. \label{q:state}Eligibility does not depend on the level of state representation. Reinforcement-learning agents over low-level state spaces were excluded (E2/E3 with I1 unmet) when the contribution was policy-learning performance rather than validation, whereas an RL or closed-loop-control paper contributing a validation, monitoring, or assurance method over trajectories satisfies I2 through closed-loop interaction.
\endgroup

The review focuses on software-level validation of agentic systems. Formal verification of hybrid systems and model checking of finite-state software provide background context, while hardware reliability and sensor fault tolerance are included only when they directly affect software-level validation claims.

\subsection{Paper Selection}
\label{sec:paper-selection}

\begin{sidewaysfigure}
  \centering
  \includegraphics[width=0.97\textheight]{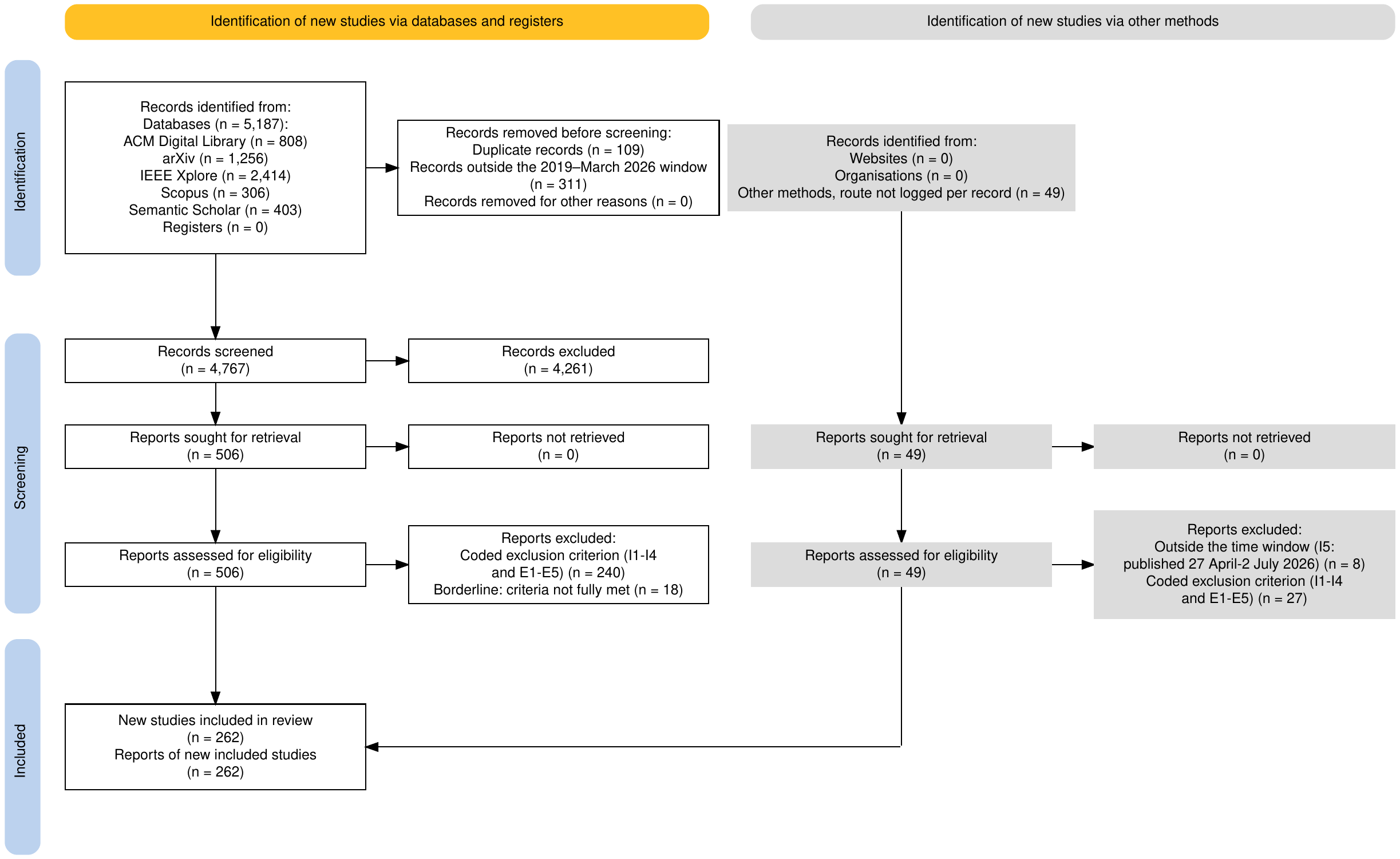}
  \caption{\add{Study-selection} workflow for literature identification, screening, eligibility assessment, and inclusion. \add{The retrieval stage returned 5{,}187 records, of which 4{,}767 were unique after removal of 311 records outside the 2019--March 2026 window and of duplicates (8 within the arXiv results, 101 across sources). Triage produced 506 candidates from this corpus; the other 49 candidates were identified through other methods (Section~\ref{sec:paper-selection}). All eight I5 exclusions are in the other-methods branch. Full-text assessment (``reports'' are papers) reduced the candidates to} \textbf{\add{262} papers} included in the final survey. \add{Diagram generated with the PRISMA2020 R package~\citep{haddaway2022prisma2020}, with box labels adapted to this review.}}
  \add{\Description{PRISMA 2020 flow diagram with two identification branches. Databases and registers: 5,187 records from ACM Digital Library (808), arXiv (1,256), IEEE Xplore (2,414), Scopus (306) and Semantic Scholar (403); 109 duplicates and 311 records outside the date window removed; 4,767 records screened and 4,261 excluded; 506 reports sought, none unretrieved, 506 assessed and 258 excluded (240 by a coded criterion I1-I4 or E1-E5, 18 borderline). Other methods: 49 records identified through other methods with no per-record route logged, 49 reports sought, none unretrieved, 49 assessed and 35 excluded (8 outside the time window under I5, published 27 April to 2 July 2026; 27 by a coded criterion I1-I4 or E1-E5). 262 new studies included.}}
  \label{fig:prisma-flow}
\end{sidewaysfigure}

\begingroup
Selection proceeded in two stages after deduplication, each applying the criteria of Section~\ref{sec:inclusion-criteria}; Figure~\ref{fig:prisma-flow} reports the record counts.

\noindent\textbf{Stage 1: Candidate triage (4,767 records to 555 candidates).} A single reviewer screened the unique records of the screening corpus against I1--I5 and E1--E5, first on title alone and then on title and abstract. Of the 555 candidates that entered eligibility assessment, 506 match a record of the screening corpus by normalized title, so 4,261 corpus records were not retained at triage. The other 49 candidates (14 of them later included) have no title match in the screening corpus. \label{q:49}They were identified outside the database searches, through forward and backward citation searching (Section~\ref{sec:search-strategy}) and other sources the authors encountered informally, such as blogs, news, and conversations with other researchers; neither the per-record triage decisions nor the source of individual candidates was logged.

\noindent\textbf{Stage 2: Candidate assessment (555 candidates to 262 papers).} Candidates were assessed in full text against all criteria, and the decision log records the decision, criterion, and rationale for every candidate. Of the 293 candidates not retained, 275 carry an explicit criterion code (240 from the screening corpus and 35 among the 49 candidates identified through other methods) and 18 are borderline records that did not fully meet the criteria, recorded with a lower-confidence note instead of a code. The criteria cited most often among the 275 coded exclusions (several may apply to one record) were I1 (142 records: contribution not validation, verification, or monitoring), E3 (102: non-agentic adjacent AI), I2 (84: no agentic execution feature), I4 (46: insufficient technical content), E1 (19), E5 (13), I5 (8: outside the time window), E4 (7), E2 (4), and I3 (2). All eight I5 exclusions are candidates identified through other methods, not records of the screening corpus: they were published between 27 April and 2 July 2026, after the end of the search window, and were excluded on the publication date alone. Of the 262 included papers, 241 were assigned high and 21 medium primary-dimension coding confidence, based on centrality to a taxonomy dimension.
\endgroup

\subsection{Coding Scheme}\label{sec:coding-scheme}

The unit of analysis is the individual paper, even when multiple systems or evaluations are reported. Each paper was coded \add{for its primary validation dimension and its approach families; study type and eight system attributes were also recorded (Section~\ref{sec:scope})}. The validation dimensions (behavioral, safety, temporal, regulatory, and multi-agent) are defined \add{conceptually in Section~\ref{sec:taxonomy} and operationally in the codebook (Section~\ref{sec:coding-rules})}. Papers could contribute evidence to more than one dimension, but a single \emph{primary} dimension was assigned based on the main validation objective emphasized by the work; secondary contributions are recorded in the coding trace but do not duplicate counts across the primary-dimension totals reported in Section~\ref{sec:approaches}.

\add{Because screening was performed by a single reviewer, agreement could be assessed only for taxonomy coding, through two checks. In the \emph{inter-rater} check, a second reviewer independently assigned a primary dimension to all 262 papers without access to the first reviewer's labels. The raters agreed on 226 papers (86.3\%; Cohen's $\kappa = 0.812$~\citep{cohen1960kappa}). The 36 disagreements concentrate at a few boundaries: behavioral/safety (12), regulatory/safety (6), behavioral/multi-agent (5), safety/multi-agent (4), temporal/regulatory (4), behavioral/regulatory (2), temporal/safety (2), and regulatory/multi-agent (1). All analyses use the first reviewer's labels. Reconciliation decisions, taken after both labels were known, are not propagated to the analysis labels; Table~\ref{tab:sens-subsets} reports the counts under the second reviewer's labels and with the reconciliation decisions applied. In the \emph{intra-rater} check, the primary reviewer applied the codebook in three passes: Pass~1 assigned a primary dimension and a primary-dimension coding confidence label (high/medium) to all 262 papers; Pass~2 re-coded, after a week-long interval, an 85-paper audit subset (all 21 papers with medium primary-dimension confidence plus a stratified sample of 64 with high confidence); Pass~3 reconciled discordant cases against the written codebook. Within-reviewer agreement was 73/85 (85.9\%) before reconciliation; the 12 discordant cases fell at five boundaries (behavioral/safety 6, safety/regulatory 3, and one each at behavioral/multi-agent, regulatory/multi-agent, and temporal/safety). Neither check covers the approach-family and secondary-dimension layer used in the coverage matrix (Section~\ref{sec:coding-rules}).}

\subsection{Coding Rules and Approach Families}
\label{sec:coding-rules}

\begingroup
Each paper receives as primary dimension the one whose validation object, as defined in Section~\ref{sec:taxonomy}, is most central to its contribution. Recurring ties at four boundaries (behavioral/safety, temporal/regulatory, safety/regulatory, and safety/multi-agent) are resolved by written borderline rules; residual ties are recorded with medium confidence and a secondary dimension. Medium confidence is also assigned when the paper's vocabulary leaves its primary validation object partially ambiguous; a secondary dimension is recorded only when the paper contributes substantively to a second dimension. The full codebook, with the operational definitions and the borderline rules, is released with the data (see Data availability).

The approach families used in the coverage matrix are defined by the kind of validation evidence a paper contributes, not by its topic (Table~\ref{tab:families}). A paper receives one tag per family whose criterion is met by a substantive contribution (a method it proposes or evidence it produces, not a family mentioned in related work); testing a proposed method on an existing benchmark does not add the Benchmarks tag. Each tag yields one appearance, placed in the column of the paper's \emph{primary} dimension; secondary dimensions are recorded but not used in the coverage matrix, and no intensity levels are coded. Papers matching no family are labeled \emph{unmatched}. This layer was coded by the primary reviewer alone, with a recorded evidence excerpt and page for each paper, and has no second-rater agreement estimate.

\begin{table}[!htbp]
\centering
\caption{Approach families used in the coverage matrix and the evidence criterion for each tag.}
\label{tab:families}
\footnotesize
\setlength{\tabcolsep}{3pt}
\renewcommand{\arraystretch}{1.1}
\begin{tabularx}{\columnwidth}{@{}>{\raggedright\arraybackslash}p{2.3cm}X@{}}
\toprule
\textbf{Family} & \textbf{Criterion (kind of evidence contributed)} \\
\midrule
Benchmarks & Scored evaluation on a fixed or refreshable suite of tasks or environments, including LLM-as-judge and process-scoring protocols; the paper builds and reports its own suite or protocol. \\\hline
Runtime Monitors & Mechanisms that observe an executing trajectory and detect, block, or contain violations within the session (guards, shields, policy enforcement, anomaly detectors). \\\hline
CPS Validation & Validation in closed loop with a physical or cyber-physical process (simulation, digital twin, hardware-in-the-loop, field), where timing and physical state determine correctness. \\\hline
Assurance Cases & Structured arguments (GSN, CAE, reusable patterns) linking claims to evidence, including automated or continuous case management. \\\hline
MLOps/Drift & Post-deployment practices that detect distribution, performance, or behavioral change and trigger revalidation, retraining, or rollback. \\
\hline
Reg Guidance & Analysis, interpretation, or operationalization of regulatory, standards, or governance obligations (FDA, MDCG, IEC 62304, liability frameworks). \\
\bottomrule
\end{tabularx}
\end{table}
\endgroup

\subsection{Corpus Composition}
\label{sec:corpus-composition}

\begin{figure}[!htbp]
\centering
\caption{Primary-dimension composition of the corpus by publication-year band\add{ (2019--2023, 2024, 2025, and January--March 2026, by the year of the recorded version; no paper predates 2019)}. Each included paper carries exactly one primary dimension, so the per-band counts ($n$) sum to the \add{262} included papers.}
\label{fig:year-dist}
\includegraphics[width=1\columnwidth]{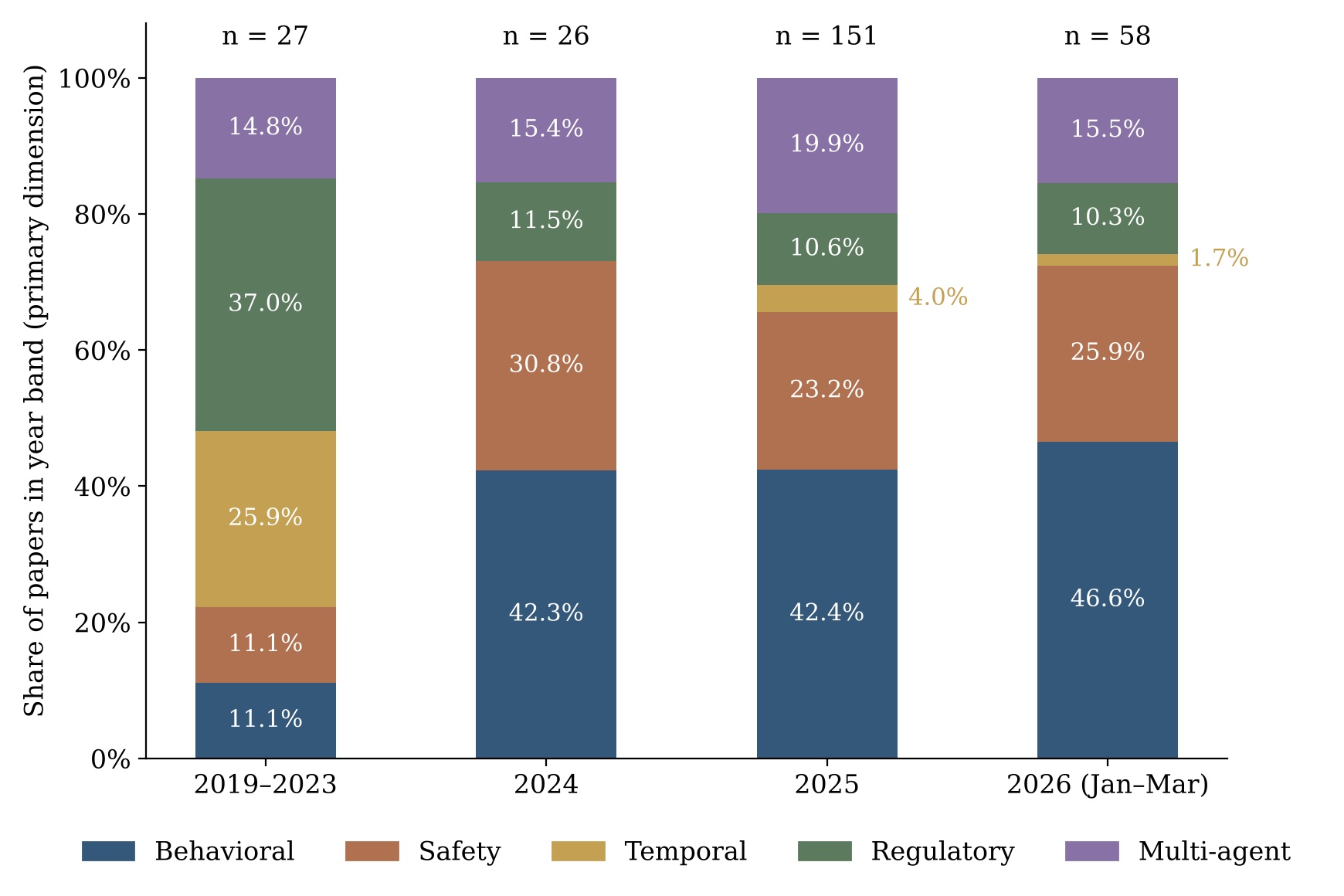}
\smallskip
\add{\Description{Stacked bar chart of primary-dimension shares for four year bands: 2019--2023 (n=27), 2024 (n=26), 2025 (n=151), and 2026 (n=58); behavioral work dominates 2024--2026, while regulatory and temporal work are largest in the earliest band.}}
\end{figure}

\add{Each included paper carries exactly one primary dimension, which yields 105 behavioral (40.1\%), 61 safety (23.3\%), 47 multi-agent (17.9\%), 35 regulatory (13.4\%), and 14 temporal (5.3\%) papers. The corpus is also heavily concentrated in recent years: no paper predates 2019, 27 date from 2019--2023, 26 from 2024, 151 from 2025, and 58 from January--March 2026, so that 209 papers (79.8\%) are from 2025 onward (Figure~\ref{fig:year-dist}). Finally, 151 papers (57.6\%) are peer-reviewed publications and 111 (42.4\%) are preprints, with a marked difference across dimensions (published share: regulatory 29/35, temporal 10/14, safety 33/61, behavioral 57/105, multi-agent 22/47). Section~\ref{sec:methodology-limits} examines how these patterns depend on publication status, source, and year.}

\subsection{Limitations and Threats to Validity}
\label{sec:methodology-limits}

\add{Two limitations concern how the corpus was assembled. First, the terminology of agentic AI validation is unstable, and terms such as ``assurance,'' ``evaluation,'' and ``monitoring'' span overlapping research programs; despite source-specific queries, citation chaining, and the inclusion of regulatory documents, terminology emerging after the March 2026 cut-off and metadata inconsistencies in preprint-heavy venues remain residual risks. Because the IEEE Xplore query uses the taxonomy's vocabulary, retrieval may also favor papers that already frame their contribution in those terms. Second, a single reviewer performed triage and full-text assessment, and no sample of excluded records was double-screened, so selection bias cannot be quantified; the inter-rater $\kappa$ concerns only the classification of the included papers. \label{q:conditional}Every corpus-level finding is therefore conditional on one reviewer's application of the criteria, and the analysis emphasizes directional patterns. The 49 candidates found outside the database searches came from a route that is neither systematic nor reproducible; the 14 papers it contributed to the corpus are spread across dimensions (multi-agent 5, safety 3, behavioral 3, regulatory 2, temporal 1), and without them the primary-dimension counts are 102/58/13/33/42 of 248 papers, with the same ordering. On the coding side, the approach-family layer has no agreement estimate (Table~\ref{tab:matrix-labelings} shows how the matrix cells cited in Section~\ref{sec:agenda} change under the second reviewer's and the reconciled labels); Table~\ref{tab:sensitivity-bounds} bounds the effect of the 12 intra-rater discordant cases.}

\begingroup
Table~\ref{tab:sens-subsets} shows how far the directional claims depend on these choices. Behavioral work is the largest category and temporal work the smallest in every row (tied with regulatory when IEEE-only papers are removed), including when only the 224 empirical studies (as labeled in the released coding sheet) are kept. Treating functional safety and adversarial security as separate dimensions (Table~\ref{tab:gap-matrix}) changes this ordering only for papers published before 2025, where adversarial security (2 papers) falls below temporal work (7). The regulatory share is not stable: it falls from 13.4\% to 3.9\% without the 110 IEEE-only papers, and 29 of its 35 papers are peer-reviewed, whereas 25 of the 47 multi-agent papers are preprints. Table~\ref{tab:ieee-cells} shows that the benchmark cells keep or increase their share without IEEE-only papers, while all CPS-validation $\times$ safety and assurance-case $\times$ temporal appearances come from IEEE-only papers, so this density reflects an IEEE-heavy retrieval set, partly driven by the broadest of the five queries. The converse bias cannot be measured, because venues such as NeurIPS, ICML, ICLR, ACL, and EMNLP were reached only through arXiv and Semantic Scholar. \label{q:recency}Recency matters as well: the 53 papers published before 2025 are spread more evenly across dimensions, so the dominance of behavioral work and the thin temporal column largely reflect the recent growth of LLM-agent evaluation, and the corpus cannot show whether these patterns are stable over time. \label{q:matrixlabels}Rater disagreement also affects the coverage matrix only modestly: under the second reviewer's or the reconciled labels, the cells cited in the research agenda change by at most two appearances, and the empty cells stay empty (Table~\ref{tab:matrix-labelings}). This check covers the assignment of papers to columns, not the approach-family tags themselves, which remain single-coder, nor screening decisions, whose error cannot be bounded with the available records.

\begin{table}[!htbp]
\centering
\caption{Primary-dimension counts (\% of subset) under alternative corpus definitions and labelings. ``IEEE-only'': all title matches from IEEE Xplore (110 papers; 14 with no match kept). ``Reconciliation applied'': first reviewer's label replaced by the logged reconciliation decision (intra-rater if the two records differ).}
\label{tab:sens-subsets}
\footnotesize
\setlength{\tabcolsep}{3pt}
\renewcommand{\arraystretch}{1.08}
\begin{tabularx}{\columnwidth}{@{}>{\raggedright\arraybackslash}p{0.24\columnwidth}>{\centering\arraybackslash}p{0.05\columnwidth}*{5}{>{\centering\arraybackslash}X}@{}}
\toprule
\textbf{Corpus / labeling} & \textbf{$n$} & \textbf{Behav.} & \textbf{Safety} & \textbf{Temp.} & \textbf{Reg.} & \textbf{Multi-ag.} \\
\midrule
Full corpus (first reviewer) & 262 & 105 (40.1) & 61 (23.3) & 14 (5.3) & 35 (13.4) & 47 (17.9) \\
Published only & 151 & 57 (37.7) & 33 (21.9) & 10 (6.6) & 29 (19.2) & 22 (14.6) \\
Preprints only & 111 & 48 (43.2) & 28 (25.2) & 4 (3.6) & 6 (5.4) & 25 (22.5) \\
High primary-dimension confidence only & 241 & 95 (39.4) & 58 (24.1) & 10 (4.1) & 33 (13.7) & 45 (18.7) \\
Excluding IEEE-only & 152 & 72 (47.4) & 35 (23.0) & 6 (3.9) & 6 (3.9) & 33 (21.7) \\
Published before 2025 & 53 & 14 (26.4) & 11 (20.8) & 7 (13.2) & 13 (24.5) & 8 (15.1) \\
Published 2025 onward & 209 & 91 (43.5) & 50 (23.9) & 7 (3.3) & 22 (10.5) & 39 (18.7) \\
Second reviewer's labels & 262 & 104 (39.7) & 69 (26.3) & 10 (3.8) & 32 (12.2) & 47 (17.9) \\
Reconciliation applied & 262 & 103 (39.3) & 61 (23.3) & 15 (5.7) & 34 (13.0) & 49 (18.7) \\
Empirical studies only & 224 & 103 (46.0) & 54 (24.1) & 7 (3.1) & 16 (7.1) & 44 (19.6) \\
\bottomrule
\end{tabularx}
\end{table}

\begin{table}[!htbp]
\centering
\caption{Selected coverage-matrix cells with and without IEEE-only papers: appearances / papers in the column's primary dimension (\%).}
\label{tab:ieee-cells}
\footnotesize
\setlength{\tabcolsep}{3pt}
\renewcommand{\arraystretch}{1.08}
\begin{tabularx}{\columnwidth}{@{}>{\raggedright\arraybackslash}X>{\raggedright\arraybackslash}X>{\centering\arraybackslash}X>{\centering\arraybackslash}X@{}}
\toprule
\textbf{Family} & \textbf{Dimension} & \textbf{Full corpus} & \textbf{Excluding IEEE-only} \\
\midrule
Benchmarks & Behavioral & 97/105 (92.4) & 67/72 (93.1) \\
Benchmarks & Multi-agent & 36/47 (76.6) & 28/33 (84.8) \\
Runtime Monitors & Safety & 32/61 (52.5) & 15/35 (42.9) \\
CPS Validation & Safety & 6/61 (9.8) & 0/35 (0.0) \\
Assurance Cases & Temporal & 8/14 (57.1) & 0/6 (0.0) \\
Assurance Cases & Regulatory & 20/35 (57.1) & 4/6 (66.7) \\
Reg Guidance & Regulatory & 13/35 (37.1) & 2/6 (33.3) \\
\bottomrule
\end{tabularx}
\end{table}

\begin{table}[!htbp]
\centering
\caption{Coverage-matrix cells cited in the research agenda (Section~\ref{sec:agenda}) and column totals under three primary-dimension labelings. The approach-family tags are unchanged (single coder); only the column to which a paper's appearances are assigned changes. ``Reconciled'': first reviewer's label replaced by the logged inter- and intra-rater reconciliation decisions, as in Table~\ref{tab:sens-subsets}.}
\label{tab:matrix-labelings}
\footnotesize
\setlength{\tabcolsep}{3pt}
\renewcommand{\arraystretch}{1.08}
\begin{tabularx}{\columnwidth}{@{}>{\raggedright\arraybackslash}X>{\centering\arraybackslash}p{0.17\columnwidth}>{\centering\arraybackslash}p{0.17\columnwidth}>{\centering\arraybackslash}p{0.17\columnwidth}@{}}
\toprule
\textbf{Cell or total} & \textbf{First reviewer} & \textbf{Second reviewer} & \textbf{Reconciled} \\
\midrule
Behavioral $\times$ Assurance Cases & 0 & 0 & 0 \\
Temporal $\times$ Benchmarks & 4 & 5 & 5 \\
Temporal $\times$ CPS Validation & 1 & 0 & 1 \\
Regulatory $\times$ Runtime Monitors & 6 & 6 & 7 \\
Multi-agent $\times$ Benchmarks & 36 & 36 & 38 \\
Multi-agent $\times$ Assurance Cases & 0 & 0 & 0 \\
Multi-agent $\times$ MLOps/Drift & 0 & 0 & 0 \\
\midrule
Appearances B/S/T/R/M & 116/81/20/52/49 & 113/95/13/48/49 & 112/83/21/51/51 \\
Papers B/S/T/R/M & 105/61/14/35/47 & 104/69/10/32/47 & 103/61/15/34/49 \\
\bottomrule
\end{tabularx}
\end{table}

\label{q:volume}These results also delimit how the counts should be read. \label{q:quality}No study-quality appraisal was performed, so a high count does not show that the methods of a dimension are mature, standardized, or strong as assurance arguments, and a low count does not show that existing methods are weak. \label{q:llm}Nor can the corpus be split by LLM status, which the coding sheet does not record: qualitatively, the behavioral and benchmark-heavy patterns mainly reflect the LLM-agent evaluation literature, whereas the CPS-validation, assurance-case, and regulatory counts come largely from IEEE-indexed work on autonomous and safety-critical systems. \label{q:refaudit}Finally, only 62 of the 214 works cited in this paper belong to the corpus; the other 152 are contextual references not counted in any statistic.
\endgroup

\begin{table}[!htbp]
\centering
\caption{Sensitivity of primary-dimension counts to adversarial reassignment of the twelve Pass~1/Pass~2 discordant cases. ``Nominal'' is the first-reviewer count; ``Range'' gives the minimum and maximum counts under least and most favorable reassignment. The 262-paper total is preserved.}
\label{tab:sensitivity-bounds}
\footnotesize
\renewcommand{\arraystretch}{1.08}
\begin{tabularx}{\columnwidth}{@{}>{\raggedright\arraybackslash}X>{\centering\arraybackslash}X>{\centering\arraybackslash}X@{}}
\toprule
\textbf{Dimension} & \textbf{Nominal} & \textbf{Worst-case range}\\
\midrule
Behavioral   & \add{105} & \add{[99, 106]}\\
Safety       & \add{61} & \add{[58, 68]}\\
Temporal     & \add{14} & \add{[13, 14]}\\
Regulatory   & \add{35} & \add{[33, 37]}\\
Multi-agent  & \add{47} & \add{[47, 49]}\\
\bottomrule
\end{tabularx}
\end{table}

\section{Related Work}
\label{sec:related}

This section positions the survey against five adjacent literatures that partially address agent validation but rarely yield a single assurance framework. The through-line is that each body of work captures one part of the problem, whether benchmarking, architectural evaluation, CPS validation, runtime assurance, or governance, while leaving other validation objects under-specified. Together, they show the need for a lifecycle account of what must be validated, when validation must be renewed, and what evidence must remain available for review.

\subsection{Classical SE Testing Limits}

Recent evaluation work converges on a familiar software-engineering lesson: task success is not equivalent to system assurance. Agent-evaluation surveys show that prevailing benchmarks privilege outcome completion while under-specifying trace quality and tool-use correctness~\citep{yehudai2025,luo2025}. Work on diagnosis makes the remaining limitation explicit: failure explanations are rarely part of the benchmark target~\citep{kddeval2025}. Static benchmark suites are vulnerable to contamination and leaderboard aging~\citep{liang2022helm, lipton2018mythos}; recent evidence documents these effects directly in agent evaluation~\citep{benchcontam2025}. AgentBench~\citep{agentbench2023} established the baseline for multi-environment evaluation but reports task completion rather than trajectory acceptability; SWE-rebench~\citep{swerebench2025} and ITBench~\citep{itbench2025} show that contamination-free refresh can materially change ranking, confirming that outcome scores are environment-snapshot artefacts. Convergent benchmark work such as AgentBoard, TheAgentCompany, and AgentHarm is important here because it broadens evaluation from static question answering to multi-turn interaction, consequential office-style tasks, and harmful-action exposure, yet still leaves intermediate-step legibility and trajectory acceptability only partially specified: AgentBoard and TheAgentCompany broaden the interaction setting~\citep{agentboard2024,theagentcompany2025}, whereas AgentHarm foregrounds harmful-action exposure~\citep{agentharm2024}. \add{Studies from several application areas point to the same gap:} clinical and cybersecurity studies report it in high-consequence settings~\citep{autonomousAgentic2025,agenticSecurity2025}, and logistics and planning studies report the same limitation in operational workflows~\citep{ablLlmbased2025,agenticPlanning2025}. The practical limit of this literature is therefore not lack of evaluation effort, but the narrow validation target encoded in most benchmark designs.

\subsection{Agentic Frameworks Evaluation Gaps}

A second literature evaluates agent frameworks directly, but in terms of capability rather than assurance. \citet{wang2024survey} provide the most comprehensive architectural survey of LLM-based agents, covering memory, planning, tool use, and interaction patterns, but do not define validation dimensions or conduct systematic corpus coding. \citet{acharya2025} and \citet{abouali2025air} offer further architectural surveys with attention to safety risks and governance, but neither treats temporal lifecycle validity or regulatory evidence legibility as first-class assurance targets. More focused work adds trajectory awareness only partially: MultiAgentBench~\citep{multiagentbench2025} and MAST~\citep{mast2025} score coordination outcomes and failure modes, AutoGen-style framework studies~\citep{autogen2025} expose orchestration patterns, recent protocol-oriented work such as Autogenesis~\citep{autogenesis2026} makes explicit that agent protocols themselves can evolve across deployment, and studies of asynchronous software-engineering agents show that handoff, parallelism, and coordination policy materially affect system behavior even when the base model is fixed~\citep{caid2026}, yet the field still lacks shared methods for validating whether trajectories remain acceptable under changing tools, humans, and environments~\citep{ luo2025,zou2025}. In other words, framework surveys explain how agents are assembled, but still leave open how those assemblies should be assured once their components evolve in deployment.

\subsection{CPS Trajectory Validation}

The CPS literature contributes a more deployment-grounded view by treating behavior as closed-loop, temporal, and environment-coupled. Work on digital twins, runtime safety, autonomy in medical and industrial systems, and International Conference on Cyber-Physical Systems (ICCPS)-style validation emphasizes that hazards emerge from feedback, delays, stale state, and coupled infrastructure, not only from wrong final outputs. \add{Digital-twin work proposes virtual replicas for the safety and security validation of autonomous driving~\citep{veledar2019dt}, and ICCPS work monitors safety at runtime under distribution shift~\citep{lin2025iccps} or verifies it quantitatively from perception~\citep{brown2025iccps}. Healthcare reviews and perspectives extend the discussion to clinical deployment~\citep{zhao2026,sharma2026}, and a scoping review finds most agentic systems in healthcare exploratory, with little clinical validation~\citep{collaco2026}; healthcare red-teaming and simulation work shows related risks~\citep{draelos2026,synthea2026},} including in gym-style evaluation environments~\citep{healthgym2025}. This literature is closest to the present paper's assurance target, but it is usually domain-specific and does not systematically connect CPS validation concerns to the broader wave of LLM-based agentic systems. Its importance here is conceptual: it shows that trajectory validity is inherently relational and time-dependent, even when the surrounding evaluation culture is not yet organized around LLM-agent terminology.

\subsection{Runtime Assurance and Monitoring}

Runtime-assurance work addresses the gap that static testing cannot close once systems adapt in the field. \citet{greshake2023} established the indirect prompt-injection problem, where malicious content in tool outputs hijacks agent behavior, \add{a failure mode on which much subsequent agent-security work builds}, motivating the trajectory-level containment gap identified in Section~\ref{sec:approaches}. Research on runtime verification, assurance monitors, drift management, policy shields, and expert monitoring shows how safety envelopes, conformance checks, and evidence refresh can intervene after deployment rather than treating release as the end of validation. Assurance and runtime-verification work supplies conformance and safety-envelope mechanisms~\citep{hawkins2021amlas,rtbas2025}, while MLOps and expert-monitoring work focuses on evidence refresh in operation~\citep{mlops2023,expertmonitoring2024}. Safe control and regulatory guidance specify complementary obligations~\citep{alshiekh2018safe,fda-pccp2025}; \add{EU post-market surveillance guidance for medical devices and anomaly monitoring extend these concerns to deployed systems}~\citep{mdcg2025-10,yourAnomaly2025}. More agent-specific systems such as AgentGuard and ControlA move from static evaluation toward runtime verification and workflow control for autonomous or semi-autonomous agents, addressing local monitoring and containment problems within an agentic execution~\citep{agentguardRuntime2025,controlaAgentic2025}. What remains missing is integration: most monitoring papers specify local safeguards, but not a full stack connecting requirements, simulation, runtime evidence, and audit-ready assurance artifacts. This is the clearest sign that runtime assurance is a necessary layer of validation, but not a substitute for the upstream and downstream evidence structures that make interventions explainable.

\subsection{Regulatory and Assurance Frameworks}

Regulatory and assurance-case literatures \add{set out the accountability expectations that technical evaluation is meant to support}. \add{Regulation, guidance, and standards from the EU, the FDA, the MDCG, the International Electrotechnical Commission (IEC), and the Institute of Electrical and Electronics Engineers (IEEE), which differ in authority, consistently foreground} lifecycle evidence, change control, documentation, human oversight, and post-market surveillance. FDA guidance foregrounds lifecycle evidence and good machine-learning practice~\citep{fda-action-plan,fda-gmlp}, while \add{its final guidance on predetermined change control plans addresses adaptive change explicitly~\citep{fda-pccp2025} and its non-binding draft lifecycle guidance extends this to lifecycle evidence}~\citep{fda-aiedsflm2025}. MDCG documents address \add{the interplay of the medical-device regulations with the AI Act and post-market surveillance}~\citep{mdcg2025-6,mdcg2025-10}; the EU MDR \add{imposes the binding post-market and documentation obligations that this non-binding guidance explains}~\citep{eu-mdr}\add{, and} IEC 62304\add{ provides the software-lifecycle standard}~\citep{iec62304}, with IEEE guidance providing an additional reference point~\citep{ieeeguide2025}. Assurance-case methods, including Goal Structuring Notation (GSN)-style argumentation and recent responsible-AI frameworks for healthcare autonomy, show how heterogeneous evidence can be assembled into reviewable claims~\citep{alelyani2025}. The unresolved problem is evidentiary generation: governance documents say what must be shown, but not how agentic systems should be tested and monitored so that those claims can be credibly made. Recent assurance-oriented work on agentic GenAI risk mapping and broader AI-safety assurance methods is therefore relevant to the present corpus because it starts to translate governance obligations into candidate evidence structures, even if the trajectory-level instrumentation problem remains open~\citep{reliabilitybydesignAgentic2025,landscapeSafety2025}.

\subsection{Positioning of This Paper}

Our paper sits at the intersection of these five axes. Relative to classical testing, it explains why component-oriented methods remain necessary but no longer define the full assurance target. Relative to agent-framework evaluation, it shifts attention from benchmark scores to trajectory validation in context. Relative to CPS work, it generalizes environment-coupled validation beyond any single domain. Relative to runtime-assurance research, it places monitoring inside a lifecycle stack rather than as an isolated safeguard. Relative to regulatory and assurance-case work, it translates governance demands into concrete validation objects. The integrative contribution is a unified view of the validation problem across the residual gaps of these literatures.

To situate this survey relative to the closest adjacent reviews, we compare them along six properties: publication year, whether they \add{rest on} a systematically screened corpus, whether they organize the field by validation objects rather than capabilities or domains, which of the five validation dimensions they treat as primary axes, whether they contain an explicit gap analysis, and whether they produce a research agenda with candidate metrics. These are the properties that most clearly distinguish an inspectable synthesis survey from a high-level overview.

\begin{table*}[!tbp]
\centering
\caption{Comparison with adjacent surveys.\\ \add{Corpus: systematically screened corpus or curated overview. Organization: the axis by which the survey structures the field. }Primary dimensions: B = Behavioral, Sa = Safety, T = Temporal, R = Regulatory, M = Multi-agent. \add{Gap analysis: explicit, partial, or absent. Agenda: research directions with or without candidate metrics, or none.}}
\label{tab:survey-comparison}
\footnotesize
\setlength{\tabcolsep}{3pt}
\renewcommand{\arraystretch}{1.08}
\begin{tabularx}{\textwidth}{@{}>{\raggedright\arraybackslash}X c l l l l l@{}}
\toprule
\textbf{Survey} & \textbf{Year} & \textbf{Corpus} & \textbf{Organization} & \textbf{Dimensions} & \textbf{Gap analysis} & \textbf{Agenda} \\
\midrule
\citeauthor{wang2024survey} & \citeyear{wang2024survey} & \add{Overview} & \add{Capabilities} & \{B\} & \add{Partial} & \add{Without metrics} \\
\midrule
\citeauthor{luo2025} & \citeyear{luo2025} & \add{Overview} & \add{Capabilities} & \{B\} & \add{Partial} & \add{Without metrics} \\
\midrule
\citeauthor{yehudai2025} & \citeyear{yehudai2025} & \add{Overview} & \add{Capabilities} & \{B\} & \add{Partial} & \add{None} \\
\midrule
\citeauthor{acharya2025} & \citeyear{acharya2025} & \add{Overview} & \add{Capabilities} & \{B\} & \add{Absent} & \add{None} \\
\midrule
\citeauthor{abouali2025air} & \citeyear{abouali2025air} & \add{Overview} & \add{Capabilities} & $\emptyset$ & \add{Absent} & \add{None} \\
\midrule
\citeauthor{zhao2026} & \citeyear{zhao2026} & \add{Systematic} & \add{Domains} & \{B, Sa, R\} & \add{Partial} & \add{Without metrics} \\
\midrule
\citeauthor{kddeval2025} & \citeyear{kddeval2025} & \add{Overview} & \add{Capabilities} & \{B, Sa\} & \add{Partial} & \add{Without metrics} \\
\midrule
\citeauthor{zou2025} & \citeyear{zou2025} & \add{Overview} & \add{Capabilities} & \{B, M\} & \add{Partial} & \add{Without metrics} \\
\midrule
\citeauthor{trustagent2025} & \citeyear{trustagent2025} & \add{Overview} & \add{Capabilities} & \{Sa, M\} & \add{Partial} & \add{Without metrics} \\
\midrule
\citeauthor{comparativeSurvey2025} & \citeyear{comparativeSurvey2025} & \add{Overview} & \add{Capabilities} & \{Sa, M\} & \add{Absent} & \add{None} \\
\midrule
\citeauthor{zha2023datacentric} & \citeyear{zha2023datacentric} & \add{Overview} & \add{Domains} & \{B\} & \add{Absent} & \add{Without metrics} \\
\midrule
\citeauthor{benchcontam2025} & \citeyear{benchcontam2025} & \add{Overview} & \add{Domains} & \{T\} & \add{Partial} & \add{Without metrics} \\
\midrule
\citeauthor{zhang2020mltesting} & \citeyear{zhang2020mltesting} & \add{Overview} & \add{Capabilities} & \{B, Sa\} & \add{Absent} & \add{Without metrics} \\
\midrule
\citeauthor{landscapeSafety2025} & \citeyear{landscapeSafety2025} & \add{Overview} & \add{Validation objects} & \{Sa, R\} & \add{Partial} & \add{With metrics} \\
\midrule
\citeauthor{collaco2026} & \citeyear{collaco2026} & \add{Systematic} & \add{Domains} & \{Sa\} & \add{Partial} & \add{Without metrics} \\
\midrule
\textbf{This survey} & \textbf{2026} & \add{Systematic} & \add{Validation objects} & \{B, Sa, T, R, M\} & \add{Explicit} & \add{With metrics} \\
\bottomrule
\end{tabularx}
\end{table*}

Table~\ref{tab:survey-comparison} shows four recurring patterns: most adjacent surveys are curated overviews rather than systematic syntheses; the field is organized by capabilities or domains rather than validation objects; temporal, regulatory, and multi-agent coverage are rarely treated as co-equal axes; and few surveys combine explicit gap analysis with a metrics-based research agenda. These are the gaps this paper is structured to address.

\section{Agentic Systems: Definitions and Validation Challenges}
\label{sec:definitions}

\subsection{What Makes Systems ``Agentic''}\label{sec:what-agentic}

We use \emph{agentic system} to denote a software system that pursues a goal through temporally extended action rather than through a single inference. Three properties matter for validation. First, the system maintains state across steps, whether through memory, retrieved context, or workflow state. Second, it acts through tool use, delegation, or other environment-coupled interventions. Third, its relevant behavior is a trajectory: planning, execution, escalation, recovery, and stopping conditions may all matter independently of the final answer. This definition covers contemporary LLM-based agents while remaining broad enough to include software-intensive autonomy in CPS and enterprise workflows. Agent-evaluation surveys support the former comparison~\citep{luo2025,zou2025}, \add{while CPS runtime-assurance studies motivate the latter~\citep{lin2025iccps,brown2025iccps}.}

\begingroup
Formally, we use \emph{trajectory} to denote a finite temporally ordered sequence
\begin{equation}
\tau=\big((s_0,a_0,m_0),\dots,(s_T,a_T,m_T)\big)\in(\mathcal{S}\times\mathcal{A}\times\mathcal{M})^{T+1},
\label{eq:trajectory}
\end{equation}
where $\mathcal{S}$ is the \emph{state space} of relevant system--environment states (possibly only partially observed, e.g., memory contents, retrieved context, tool and environment variables), $\mathcal{A}$ is the \emph{action space} (tool invocations, control commands, messages, natural-language outputs, delegation, and stopping), and $\mathcal{M}$ is the \emph{annotation-metadata space} (tool identifiers, timestamps, evidence provenance and version tags, delegation and handoff records). Each $s_t\in\mathcal{S}$ is the state at step $t$, each $a_t\in\mathcal{A}$ is the action selected at that step, and each $m_t\in\mathcal{M}$ records minimal execution annotations such as tool invocation, retrieved evidence, delegation, or human handoff.
\endgroup
 This definition is intentionally lightweight: the state may be partially observed, the action may be symbolic or natural-language, and the annotations need only be rich enough to reconstruct why a step occurred and what external intervention it triggered.

We separate three boundary terms. An \emph{agentic execution feature} is any trajectory-relevant property used for screening: planning across steps, tool invocation with feedback, persistent state, closed-loop external interaction, or coordination among autonomous components. A \emph{canonical agentic system} satisfies the three conceptual properties together: persistent state, environment-coupled action, and trajectory-level behavior. An LLM-based agent with planning loops, persistent memory, and tool-call sequences exemplifies this canonical case. A \emph{consequential assurance trigger} arises when any agentic execution feature affects workflows, infrastructures, or physical processes strongly enough that lifecycle validation becomes necessary.

This distinction matters operationally. The corpus-inclusion criterion (I2) is disjunctive because a paper enters the corpus if it addresses trajectory-level validation of any system exhibiting at least one agentic execution feature. A stateful dialog manager with no external tool calls occupies a boundary zone and is included only if the paper demonstrates that its lifecycle or trajectory-level properties matter for assurance. Conversely, a stateless tool dispatcher that invokes an external service is included if the paper treats the multi-step outcome as the unit of validation rather than only the final response. The canonical cases form the primary target of the taxonomy\add{ (corpus-level findings refer to the whole corpus; Section~\ref{sec:inclusion-criteria})}, while the permissive inclusion boundary captures consequential assurance triggers that do not emphasize all three canonical properties.

\subsection{Architecture Typology}

For validation purposes, three architectural patterns matter. \emph{Pipeline} systems follow predetermined stages with limited branching, so the main risks are interface mismatches, stale handoffs, and accumulation of stage-local errors. A clinical documentation pipeline that extracts findings, classifies urgency, and drafts a note in fixed sequence illustrates this pattern: errors in the extraction stage propagate silently into later stages, and the fixed stage order offers no mechanism for dynamic error recovery. \emph{Network} systems route dynamically across tools, planners, or peer agents, shifting attention to coordination failures, non-local causality, and weak trace reproducibility. A code-repair agent that queries a retrieval index, invokes a linter, calls an LLM planner, and conditionally delegates to a test executor illustrates this pattern: the same defect report can trigger different tool invocation sequences across runs, making it difficult to define a canonical trajectory against which to evaluate correctness. \emph{Human-in-the-loop} systems insert review, override, or escalation checkpoints, making calibration of handoffs, authority boundaries, and accountability for mixed human–agent trajectories the dominant assurance concern. A triage agent that surfaces high-risk cases to a clinician while autonomously closing low-risk ones illustrates this pattern: validation must address not only whether the agent acts correctly in isolation, but whether its escalation threshold is well-calibrated, neither suppressing alerts that require human review nor generating so many interruptions that human oversight becomes ineffective in practice.

\subsection{The Expanded Assurance Target}

Once a system is agentic, the assurance target expands from isolated component outputs to \emph{acceptable trajectories in context}. The relevant questions become whether the system gathered the right evidence, used tools appropriately, respected safety and escalation boundaries, coordinated correctly with humans or peer agents, and remained valid as the environment changed. This is why the paper's central validation object is not the model response alone, but the trajectory realized by the socio-technical system.

This expansion has a direct methodological consequence: test adequacy can no longer be defined by coverage of input–output pairs. A test suite that exercises all intended tool calls in isolation may still fail to expose trajectory-level failures, such as incorrect tool ordering, unsafe intermediate states, or recovery breakdowns, that only become visible when the full action sequence is examined end to end. The five-dimension taxonomy in Section~\ref{sec:taxonomy} articulates what
trajectory-level adequacy requires across behavioral, safety, temporal, regulatory, and multi-agent concerns.

\subsection{Scope and Boundary Conditions}\label{sec:scope}

The paper focuses on software-intensive agentic systems whose actions are consequential because they affect workflows, infrastructures, or physical processes. Full lifecycle assurance applies when systems exhibit an agentic execution feature and act consequentially; in these settings, classical testing no longer covers the full assurance target. Table~\ref{tab:intro-mismatch} and the five-dimension taxonomy in Section~\ref{sec:taxonomy} specify where that coverage breaks. The dividing line is not model size or interface type, but whether the system's assurance target requires reasoning about sequences of actions, state transitions, and environment feedback rather than about isolated responses.

These trajectory-level assurance burdens are triggered disjunctively: any one of them can be sufficient to create a validation gap that classical component testing cannot close. That is why a system that exhibits only one of the three canonical agentic properties (persistent state, environment-coupled action, trajectory-level behavior) can still warrant lifecycle assurance if it operates consequentially. The inclusion boundary (I2) therefore remains disjunctive at the corpus level, while the conceptual core of agentic systems remains the canonical case where all three properties hold. The research agenda\add{ focuses} primarily on that canonical case\add{, whereas the gap analysis covers the whole corpus}, because that is where the convergence of all five validation dimensions becomes most pressing.

\begingroup
To characterize the included systems, the coding sheet records study type and eight system attributes for the 262 included papers, each with an evidence note and a confidence label (Table~\ref{tab:system-attributes}). These descriptive counts have no independent agreement estimate and characterize the included studies, not the prevalence of system types in the wider field.

\begin{table*}[!tbp]
\centering
\caption{System attributes and study type of the 262 included papers, as recorded in the coding sheet (percentages of 262). ``n/a'' is the sheet value and does not distinguish unavailable from inapplicable information. The attributes have no independent agreement estimate; the sheet's own confidence label is high for 141 papers, medium for 118, and low for 3.}
\label{tab:system-attributes}
\footnotesize
\setlength{\tabcolsep}{4pt}
\renewcommand{\arraystretch}{1.05}
\begin{tabularx}{\textwidth}{@{}>{\raggedright\arraybackslash}X>{\raggedright\arraybackslash}Xrrc>{\raggedright\arraybackslash}X>{\raggedright\arraybackslash}Xrr@{}}
\toprule
\textbf{Attribute} & \textbf{Category} & \textbf{$n$} & \textbf{\%} & \phantom{x} & \textbf{Attribute} & \textbf{Category} & \textbf{$n$} & \textbf{\%} \\
\cmidrule(r){1-4}\cmidrule(l){6-9}
Number of agents & Multiple & 125 & 47.7 & & Tool use & Yes & 171 & 65.3 \\
 & Single & 124 & 47.3 & &  & No & 48 & 18.3 \\
 & n/a & 13 & 5.0 & &  & n/a & 43 & 16.4 \\
\addlinespace
Memory & Session & 150 & 57.3 & & Planning & Implicit & 106 & 40.5 \\
 & Persistent & 40 & 15.3 & &  & Explicit & 88 & 33.6 \\
 & None & 16 & 6.1 & &  & None & 28 & 10.7 \\
 & n/a & 56 & 21.4 & &  & n/a & 40 & 15.3 \\
\addlinespace
Environment & Simulated & 175 & 66.8 & & Human involvement & None & 190 & 72.5 \\
 & Real & 54 & 20.6 & &  & In the loop & 33 & 12.6 \\
 & None & 27 & 10.3 & &  & On the loop & 8 & 3.1 \\
 & n/a & 6 & 2.3 & &  & n/a & 31 & 11.8 \\
\addlinespace
Adaptation & None & 209 & 79.8 & & Risk level (coded) & Low & 105 & 40.1 \\
 & Self-evolving & 24 & 9.2 & &  & Medium & 85 & 32.4 \\
 & Updates & 20 & 7.6 & &  & High & 69 & 26.3 \\
 & n/a & 9 & 3.4 & &  & n/a & 3 & 1.1 \\
\midrule
\multicolumn{9}{@{}l}{\textbf{Study type:} empirical 224 (85.5\%), conceptual 31 (11.8\%), survey 4 (1.5\%), guidance 3 (1.1\%)} \\
\bottomrule
\end{tabularx}
\end{table*}
\endgroup

\section{Why Classical SE Testing Assumptions Break}
\label{sec:classical}

\subsection{Core Assumptions of Classical Testing}\label{sec:core-assumptions}

Classical software testing assumes that behavior is sufficiently stable, decomposable, specifiable, and environment-bounded that evidence from unit, integration, regression, and system tests can be composed into a credible judgment about the deployed system. Those assumptions remain valuable for deterministic subsystems, tool wrappers, validators, interfaces, and change control. They stop being sufficient when the system's relevant behavior is an open-ended action trajectory.

Each assumption encodes a tacit decomposability claim. \add{Unit testing is typically justified by the expectation that function-level correctness, together with integration evidence, supports system-level claims, although classical practice does not treat this inference as automatic.} \add{Integration testing presupposes interface contracts stable enough that interactions verified at release remain valid until a change triggers re-verification.} Regression testing assumes that prior behavior is a meaningful baseline against which to detect degradation. Specification-based testing assumes that requirements can be expressed as checkable pre- and postconditions. System testing assumes that a bounded set of representative scenarios covers the space of safety-relevant behaviors. Fault injection typically assumes that failures can be injected at identified components and traced through sufficiently bounded interactions. In agentic systems, \add{each of these assumptions is weakened}, as the structural argument below and the empirical examples below both show.

\subsection{Empirical Mismatch Examples}\label{sec:mismatch-examples}

Table~\ref{tab:comparison} summarizes the structural mismatch by operationalizing the five abstract mismatches from Table~\ref{tab:intro-mismatch} across six familiar testing methods; the fault-injection row is separated as a method-level view of breakdowns that cut across decomposition and environment mismatch rather than as a sixth abstract mismatch. \add{The examples below show that the mismatch is documented in applied studies, although trajectory-level evidence from deployed agents remains scarce.}

\begin{table*}[!tbp]
\hbadness=10000
\centering
\caption{Why classical testing abstractions are insufficient for agentic systems. The six method rows operationalize the five abstract mismatches in Table~\ref{tab:intro-mismatch}; the table is an author synthesis of the literatures discussed in this section and in Section~\ref{sec:definitions}.}
\label{tab:comparison}
\footnotesize
\renewcommand{\arraystretch}{1.08}
\begin{tabularx}{\textwidth}{@{}p{2.1cm}p{2.8cm}Xp{4.3cm}@{}}
\toprule
\textbf{Testing method} & \textbf{Core assumption} & \textbf{Why agentic systems break it} & \textbf{Operational implication} \\
\midrule
Unit testing & Reproducible function behavior & The same apparent state can induce different reasoning traces, tool calls, and memory updates & Passing micro-tests does not imply dependable autonomous behavior \\
\midrule
Integration testing & Stable interfaces and predictable composition & Tool APIs, retrieval behavior, and orchestration policies evolve during deployment & Workflow-dependent failures appear after release \\
\midrule
Regression testing & Prior behavior is a stable baseline & Memory, model updates, and environment change invalidate historical baselines & Evidence decays unless refreshed \\
\midrule
Specification-based testing & Requirements can be stated as complete properties & Goals such as ``escalate appropriately'' and ``use tools safely'' are partial and contextual & Runtime constraints and scenario evidence become necessary \\
\midrule
System testing & Representative end-to-end scenarios can be bounded & Context spans tools, humans, timing, and coordination effects that labs under-approximate & Safety-critical hazards escape fixed test suites \\
\midrule
Fault injection & Injected faults can be isolated to bounded components and propagated through tractable interactions & Breakdowns can cascade across planner, tool, memory, and human couplings and become only partially attributable at component level & Root-cause analysis becomes trajectory-level and often requires cascade-aware tracing \\
\bottomrule
\end{tabularx}
\end{table*}
Three short examples make the mismatch concrete. \add{In smart-mobility settings, runtime monitors evaluate safety over predicted future trajectories rather than single control outputs, because a collision depends on how the state evolves over a horizon~\citep{brown2025iccps,lin2025iccps}; validating the final maneuver alone misses the path-dependent hazard. In industrial systems, learned models that perform well on historical data degrade when sensors shift, equipment degrades, or operating modes change~\citep{boundaryawareConcept2025,connectionBetween2025}; for an agent acting on such models, stale telemetry, delayed actuation, or tool disagreement would further shift the safe operating window.} In clinical workflows, an agent may produce a plausible recommendation while relying on stale lab data, mis-timed escalation, or incomplete chart retrieval; the clinically salient failure is the trajectory, not the surface plausibility of the endpoint. \add{Physician red-teaming documents the response-level version of this failure, including missing escalation~\citep{draelos2026}, while trajectory-level evidence from deployed clinical agents remains scarce~\citep{collaco2026}; in MLOps, expert-monitoring work proposes routing drift signals to on-call staff together with domain knowledge of drift-inducing events~\citep{expertmonitoring2024}.}

\begingroup
\label{q:specific}Several of these limitations are not new, however. Closed-loop validation, hybrid-systems verification, runtime assurance, and fault injection (Section~\ref{sec:related}) already address environment coupling, timing, and stale state, and testing research has long dealt with nondeterminism in concurrent and distributed software. What is specific, or specifically aggravated, in agentic systems are three properties: (i) \emph{open-ended action spaces with stochastic, contingent planning}, so that the set of admissible trajectories cannot be enumerated and oracles are necessarily partial; (ii) \emph{natural-language and tool-semantic interfaces} that act as de facto contracts without formal specification and can change silently; and (iii) \emph{mutable memory} that makes behavior history-dependent and lets adaptation accumulate inside a single deployment. The determinacy and specification mismatches of Table~\ref{tab:intro-mismatch} are most directly tied to (i) and (ii), and the temporal mismatch to (iii); the environment mismatch is largely shared with CPS practice.
\endgroup

\subsection{What Still Carries Over}

Classical methods remain essential: component testing, interface contracts, hazard analysis, fault injection, and change control retain their roles, while machine-learning (ML) engineering adds versioning and conservative deployment discipline~\citep{zinkevich2016rules,zhang2020mltesting}. Testing research provides further methods for the transition~\citep{riccio2020testing,braiek2020testing}. What changes is their role: each method retains validity within a circumscribed scope, component tests for input–output-stable subsystems, contract tests at API boundaries, Failure Mode and Effects Analysis (FMEA)/Hazard and Operability Study (HAZOP) when scoped to specific tool invocations or handoff points, fault injection targeting tool-call responses and inter-agent messages, and change control for model checkpoints and tool-schema versions. What none of these supplies alone is the composition of component-level assurances into trajectory-level claims. That gap is the entry point for the taxonomy in Section~\ref{sec:taxonomy}.

\section{A Taxonomy of Validation Dimensions}
\label{sec:taxonomy}

We organize the validation problem into five dimensions derived\add{, in a hybrid deductive-then-inductive process (Section~\ref{sec:review-design}),} from \add{five assurance burdens identified in the assurance, cyber-physical, and regulatory literatures and refined in the pilot coding}: trajectory quality, bounded safety, persistence of evidence over time, regulatory legibility, and collective behavior under interaction. This taxonomy of \emph{validation objects} defines what is validated and why; coverage across the literature is examined in Section~\ref{sec:approaches}. Table~\ref{tab:taxonomy} summarizes the dimensions, their characteristic failure modes, and representative metrics.

The five dimensions correspond to five recurring questions in the surveyed literature: whether the trajectory is acceptable, whether it stays within risk bounds, whether that judgment remains valid after change, whether the supporting evidence is auditable, and whether local policies compose into acceptable joint behavior. These questions are analytically distinct even when they co-occur in deployment. A system can be behaviorally effective yet unsafe, temporally stale yet still presently auditable, or individually well-behaved yet collectively hazardous. This is why the taxonomy is organized by failure mode rather than by capability list.

\begin{table*}[!tbp]
\caption{Standalone taxonomy of validation dimensions for agentic systems. Each row states a validation object, a characteristic failure mode, general metrics, and a brief example.}
\label{tab:taxonomy}
\centering
\footnotesize
\setlength{\tabcolsep}{3pt}
\renewcommand{\arraystretch}{1.05}
\begin{tabularx}{\textwidth}{@{}>{\raggedright\arraybackslash}p{1.9cm}>{\raggedright\arraybackslash}X>{\raggedright\arraybackslash}X>{\raggedright\arraybackslash}X@{}}
\toprule
\textbf{Dimension} & \textbf{Conceptual definition (validation object + failure mode)} & \textbf{General measurable metrics} & \textbf{Brief example} \\\midrule
Behavioral & Task-level trajectory: action choice, tool sequence, state updates, and recovery under realistic variation. Failure appears as brittle or unjustified behavior across semantically similar situations. & Task-completion rate; constraint satisfaction; trajectory consistency; recovery success; tool-selection precision/recall; escalation appropriateness. & A support agent chooses different refund procedures for equivalent complaints, producing inconsistent service outcomes. \\\midrule
Safety & Risk boundary for autonomy during error, uncertainty, or interface failure. Failure appears as harmful action, unsafe tool use, or uncontrolled degradation despite nominal competence. & Unsafe-action rate; near-miss rate; severity-weighted incidents; policy violations; time to safe stop; intervention success. & A warehouse robot agent routes forklifts into a restricted area after receiving corrupted sensor data.\\\midrule
Temporal & Persistence of assurance claims as inputs, tools, policies, and memory evolve. Failure appears as evidence decay after drift, updates, or accumulated history. & Drift slope; calibration drift; memory contamination; evidence freshness; revalidation interval; post-update regression. & A rehabilitation-planning agent becomes less reliable after therapy-slot encodings change. \\\midrule
Regulatory & Evidence package for traceable, auditable, and governable behavior under external oversight. Failure appears as missing justification for intended use, change control, provenance, or accountability. & Trace completeness; audit-log coverage; requirement-to-test linkage; provenance completeness; change-impact latency; evidence-backed claims. & A loan-processing agent cannot reconstruct why it changed a credit recommendation after a model update.\\\midrule
Multi-agent & Collective behavior of interacting agents, services, and humans. Failure appears when locally acceptable policies combine into conflict, oscillation, duplication, deadlock, or unsafe delay. & Coordination success; conflict frequency; deadlock/livelock incidence; duplicated actions; interaction latency; resource contention. & A delivery-routing agent and an inventory agent issue conflicting shipment instructions, causing delays. \\\bottomrule
\end{tabularx}
\end{table*}

\begingroup
\label{q:conceptual}Conceptually, the dimensions are complementary assurance \emph{facets}, neither mutually exclusive categories nor a hierarchy, and they sit at different levels of abstraction: behavioral and safety are properties of system behavior, temporal validity concerns the persistence of a claim over the lifecycle, regulatory legibility concerns evidentiary governance, and multi-agent validity concerns interaction structure. They are parallel only in a failure-mode sense: each names a distinct way in which a trajectory-level assurance claim can fail or go unsupported, and each calls for different evidence. A paper may contribute to several facets; the single primary label is a counting device for paper-level statistics (Section~\ref{sec:coding-rules}). \label{q:codability}The taxonomy has been checked for codability (inter-rater $\kappa = 0.812$ on the primary label) but not for completeness or independence, which would require an expert panel or a formal ontology analysis.

Table~\ref{tab:taxonomy-frameworks} relates the taxonomy to established frameworks. It is meant as an overlay: the frameworks supply methods for individual facets, and the taxonomy supplies a common vocabulary for the evidence an agentic system needs across facets.

\begin{table*}[!tbp]
\centering
\caption{Relation of the taxonomy to established assurance frameworks. The mapping is an author synthesis (conceptual layer), not an empirical result.}
\label{tab:taxonomy-frameworks}
\footnotesize
\setlength{\tabcolsep}{3pt}
\renewcommand{\arraystretch}{1.1}
\begin{tabularx}{\textwidth}{@{}>{\raggedright\arraybackslash}p{2.9cm}>{\raggedright\arraybackslash}p{2.6cm}>{\raggedright\arraybackslash}X>{\raggedright\arraybackslash}X@{}}
\toprule
\textbf{Framework} & \textbf{Closest dimensions} & \textbf{What the framework supplies} & \textbf{What the taxonomy adds / does not replace} \\
\midrule
Runtime assurance and Simplex-style architectures~\citep{sha2001simplex,rushby2008runtimecert,bartocci2018specmonitoring,alshiekh2018safe} & Safety; part of Temporal & Verified fallback controller with a switching condition; specification-based monitoring of traces & Adds history-dependent, natural-language action spaces and evidence-expiry conditions; does not replace verified fallback logic \\
\midrule
Safety cases, GSN/CAE, Assurance 2.0~\citep{gsn2021,bloomfield2020assurance2} & Regulatory (format cuts across all five) & Structured claim--argument--evidence packaging and defeater handling & Names the trajectory-level claims and evidence types to be packaged; does not replace the argument structure \\
\midrule
STPA and hazard analysis~\citep{leveson2011stpa} & Safety; Multi-agent (control structure) & Hazard and unsafe-control-action identification from the control structure & Adds dimensions beyond hazards (freshness, accountability); STPA remains the source of hazards and requirements \\
\midrule
Software verification and testing~\citep{ammann2016} & Behavioral; Safety & Specification-based testing and verification for bounded programs & Marks where bounded-program assumptions fail (open-ended actions, partial oracles) \\
\midrule
ML lifecycle governance~\citep{amershi2019se4ml,sculley2015hidden,nistairmf2023,isoiec42001} & Temporal; Regulatory & Lifecycle practices, risk management, and management-system requirements & Specifies agent-level change channels (memory, tools, workflow) that trigger revalidation \\
\midrule
Multi-agent verification~\citep{dennis2020verifiable,verifyingMulti2025} & Multi-agent; Safety & Model checking and verification of agent programs and protocols & Targets open-vocabulary LLM coordination where finite-program assumptions do not hold \\
\bottomrule
\end{tabularx}
\end{table*}
\endgroup

\subsection{Behavioral Dimension}

The behavioral dimension evaluates whether an agent's realized trajectory is competent, consistent, and justifiable rather than merely whether the final answer looks plausible. The literature shows this from several directions: evaluation surveys establish the gap between outcome scores and trajectory quality~\citep{yehudai2025,luo2025}; failure-mode studies locate breakdowns in reasoning transitions and tool-invocation sequences rather than in final answers~\citep{mast2025}; data-centric work shows how quality and provenance defects propagate through otherwise plausible traces~\citep{zha2023datacentric,retrievalAugmented2025}. Applied pipeline studies confirm the same pattern in production-like settings: orchestration and simulation-optimization studies expose it in automated decision pipelines~\citep{auraAgent2025,agenticSimheuristic2025}, while workflow-generation and monitoring studies observe it in evolving processes~\citep{autoflowgenMultiagent2025,thresholdtriggeredDeep2025}. Domain-adapted deployments supply further evidence~\citep{prismaiDualstage2025,vtsllmDomainadaptive2025}. The characteristic gap is outcome-preserving but procedurally defective behavior.

The practical implication is that behavioral validation cannot be reduced to measuring task success rates at the endpoint. An agent that achieves the correct final output via an inconsistent, unjustified, or brittle trajectory offers weaker assurance guarantees than one whose intermediate steps are themselves verifiable. This matters most when intermediate tool calls produce external side effects, including database writes, API calls that trigger downstream processes, and escalation decisions that route work to human reviewers, because a procedurally defective trajectory may leave the environment in an inconsistent state even when the final answer appears correct. Trajectory consistency under repeated trials and recovery success after perturbation are therefore first-class behavioral metrics, not secondary proxies, for this dimension.

\subsection{Safety Dimension}\label{sec:safety-dim}

The safety dimension evaluates bounded autonomy: whether the system can avoid harmful actions, unsafe tool use, and uncontrolled degradation under error, uncertainty, or attack. In this survey, Safety is an intentionally aggregated dimension because functional safety and adversarial security both define the action boundary that keeps agent autonomy within acceptable risk limits. Foundational safety work and agent-specific safeguards articulate bounded action~\citep{amodei2016concrete,agentguard2025}, whereas adversarial benchmarks and autonomy analyses expose ways those bounds can be crossed~\citep{nemesisAdversarial2025,autonomousAgents20252}. Threat-adaptation and observability studies add operational evidence~\citep{adaptiveThreat2025,breakingObservability2025}, including multilingual and cross-agent attacks~\citep{adversarialMultilingual2025,crossagentMultimodal2025}. The former emphasizes hazard analysis, fail-safe behavior, bounded degradation, and safe-stop or handoff mechanisms; the latter emphasizes prompt injection, tool misuse, privilege escalation, data exfiltration, and adversarial manipulation. \add{\label{q:safetyjust}Coding the 61 safety papers by sub-strand gives 23 functional-safety, 29 adversarial-security, and 9 papers contributing to both (Table~\ref{tab:gap-matrix}), and the two strands are reported separately. They are kept in one dimension because in agentic systems the boundaries interact: a prompt injection that hijacks a tool call can initiate a hazard in a cyber-physical process, and a safe-stop or degradation policy must hold under adversarial input. The strands nevertheless differ in threat model, assurance argument, standards, and methods (hazard analysis, barrier certificates, and fallback envelopes versus semantic filtering, prompt isolation, and privilege control), so coverage in one is not coverage in the other, and we do not assume that prompt-injection defenses compose with graceful-degradation protocols (Section~\ref{sec:agenda}).} The underspecified-goal problem, in which agents appear safe while exploiting loopholes in vaguely specified objectives, is documented empirically as specification gaming and is a structural precursor to both safe-stop failures and security policy bypass~\citep{krakovna2020gaming}. The recurring gap is therefore not just hazard detection, but the specification of safe-stop, defer, or handoff behavior that remains robust under underspecified goals, partial observability, and adversarial interference.

\subsection{Temporal Dimension}\label{sec:temporal-dim}

Temporal validation asks whether assurance evidence persists after updates, drift, memory accumulation, workflow change, or benchmark aging. This dimension synthesizes work from dataset shift, MLOps, post-deployment monitoring, and production ML research, all of which show that validation is a moving target rather than a one-time pre-release event. Benchmark-contamination and MLOps studies show that evaluation conditions and models both change over time~\citep{benchcontam2025,mlops2023}; expert-monitoring and dataset-shift work describe the resulting evidence-maintenance burden~\citep{expertmonitoring2024,quionero2009dataset}. Production-ML research supplies the complementary account of hidden technical debt and operational regression~\citep{sculley2015hidden,breck2017ml}, including rules for dependable deployment~\citep{zinkevich2016rules}. Its central failure mode is evidence decay: a system once judged acceptable no longer merits the same claim after its operating context changes.

This dimension is especially consequential for agentic systems because many of their assurance claims are path-dependent rather than purely input–output based. A benchmark score or pre-deployment test result may look stable even while the underlying trajectory-generating process has changed: memory may accumulate stale assumptions, retrieved evidence may shift in quality or provenance, tool interfaces may silently change their semantics, and escalation behavior may drift as surrounding workflows adapt. Three channels recur: memory or retrieved-context drift, tool-version and schema change, and environment or workload shift. In each case, a release-time claim can become stale without the system ever failing a release-time test.

Temporal adequacy therefore requires explicit evidence-expiry conditions: what kinds of change invalidate prior claims, which traces or monitoring signals count as sufficient warning, and what revalidation scope follows. Without that linkage, monitoring remains observational rather than normative, able to report change, but not to determine when a previously accepted deployment is no longer adequately validated. Section~\ref{sec:agenda} develops this as a lifecycle-assurance direction.

\begingroup
\label{q:temporal}Five notions recur in this dimension and are related as follows. \emph{Concept or data drift}, \emph{memory contamination}, \emph{tool and interface change}, and \emph{benchmark aging} are change \emph{sources}, each acting on a different part of the system (input distribution, accumulated state, interface semantics, and evaluation suite, respectively). \emph{Evidence freshness} is the \emph{consequence for the assurance claim}: the age and continued validity of the evidence relative to the claim it supports. Freshness is necessary but not sufficient, since recent data can still be invalid. We keep the five within one dimension because a revalidation trigger is defined on the claim, not on the source of change. The corpus was not coded at the level of these sub-facets.
\endgroup

\subsection{Regulatory Dimension}\label{sec:regulatory-dim}

The regulatory dimension evaluates whether validation evidence is traceable, reviewable, and governable under external oversight. This is where audit trails, provenance, requirement-to-evidence linkage, change control, and post-market processes become part of the validation target rather than administrative afterthoughts. FDA documents articulate lifecycle evidence and good machine-learning practice~\citep{fda-action-plan,fda-gmlp}, and FDA's \add{final guidance on predetermined change control plans makes change control central~\citep{fda-pccp2025}, while its lifecycle guidance for AI-enabled device software is still a non-binding draft}~\citep{fda-aiedsflm2025}. \add{In the EU, the MDR imposes binding post-market obligations~\citep{eu-mdr}, which non-binding MDCG guidance explains, including their interplay with the AI Act}~\citep{mdcg2025-6,mdcg2025-10}, \add{and IEC 62304 situates them in software-lifecycle practice}~\citep{iec62304}, with IEEE guidance adding a further reference point~\citep{ieeeguide2025}. Assurance and evidence-design studies show how these requirements can be made reviewable~\citep{alelyani2025,assurancedrivenDesign2025}, with concise agentic proposals extending the argument~\citep{shortPaper2025}. \add{The concern this dimension captures is that even technically strong systems may lack} an evidence package that can be reconstructed and defended after deployment changes or incidents.

This gap has a structural cause: regulatory evidence requirements are typically defined at system release, but agentic systems continue to evolve as model updates, tool schema changes, retrieval corpus refreshes, and policy revisions accumulate after the initial evidence package is assembled. Maintaining regulatory legibility therefore requires not only that the release-time evidence is complete and traceable, but also that a change-impact assessment process exists to determine when a post-release change is material enough to require evidence re-submission or re-evaluation. \add{FDA guidance addresses this through predetermined change control plans for AI-enabled devices~\citep{fda-pccp2025}, and the MDCG/AIB guidance discusses how changes to AI-based device software are assessed under the MDR and the AI Act~\citep{mdcg2025-6}; how technical validation methods can support such change-impact assessment for agentic systems remains an open question.} The implication for this dimension is that audit-log coverage and provenance completeness must be treated as runtime properties maintained continuously, not as documentation artifacts assembled at release and left static thereafter.

\begingroup
\label{q:regnotions}The regulatory dimension should be distinguished from several neighboring notions. An \emph{assurance case} is an argument format; \emph{auditability} is a technical property of traces that permits reconstruction of what happened; \emph{accountability} is the allocation of responsibility among people and organizations; \emph{documentation} is a set of static artifacts; \emph{lifecycle governance} is the process that controls change. The regulatory dimension concerns the validation object that contains them: whether the evidence package as a whole can be reviewed and defended before an external party. The others are instruments or properties serving it. Consistently with the coding rules, a paper whose central output is a structured assurance argument or audit artifact is coded regulatory even when its content concerns safety mechanisms.
\endgroup

\subsection{Multi-Agent Dimension}

The multi-agent dimension evaluates the joint trajectory produced by interacting agents, services, and humans. It is structurally distinct from single-agent validation because acceptable local policies can still compose into conflict, deadlock, duplication, delay, or distributed risk. Benchmark and failure-mode evidence documents these compositional breakdowns directly~\citep{multiagentbench2025}; work on cooperation and coordination mechanisms shows how they arise from interaction structure rather than from agent-level defects. Collaborative and mean-field analyses relate failures to interaction structure~\citep{collaborativeMultiagent2025,meanfieldDeep2025}; cooperative learning and strategic-interaction work identifies further coordination mechanisms~\citep{efficientCooperative2025,strategicLearning2025}, including explicit coordination models~\citep{multiagentCoordination2025}. Studies of LLM-based frameworks and role design confirm the pattern in contemporary stacks~\citep{frameworkbasedMultiagent2025,multiagentLlm2025}, at increasing organizational scale~\citep{harnessingMultiagent2025,analysingRole2025}, and in scientific multi-agent settings~\citep{scienceScaling2025}. The joint trajectory, not any component in isolation, is therefore the proper validation object.

This makes the dimension asymmetric with the other four in a basic sense: behavioral, safety, temporal, and regulatory concerns all apply to single-agent systems as well, whereas multi-agent validation exists only once assurance claims depend on relations among multiple decision makers.
The distinctive burden here is therefore not merely to add more agents to an existing validation frame, but to validate interaction-level properties, including coordination, delegation, conflict resolution, and joint accountability, that do not arise in isolated-agent settings.

The structural challenge is that emergent coordination failures cannot generally be predicted from single-agent test results: an agent whose individual behavior is within specification may still produce unsafe outcomes when coupled with other agents whose policies interact with its own in unanticipated ways. Shared resource contention, implicit assumptions about message ordering, and uncoordinated escalation are \add{typical examples of such interactions}. Validating this dimension therefore requires test scenarios that exercise inter-agent interactions explicitly, including adversarial compositions in which one agent's well-intentioned action is precisely the trigger that causes another's policy to fail. Blame resolution and causal attribution in such settings are particularly difficult because the observable failure may occur far removed from its initiating cause, both temporally and across a chain of interacting agents.

\subsection{Dimension Intersections and Relation to Prior Surveys}

The most consequential failures often lie at dimension intersections: behavioral $\times$ safety yields competent-but-dangerous trajectories; temporal $\times$ regulatory yields stale but still-claimed evidence; and multi-agent interaction amplifies both behavioral and safety hazards through coordination breakdowns~\citep{sambasivan2021cascades}. For behavioral $\times$ safety, a concrete failure mode is an agent that competently completes its assigned task while choosing an unsafe tool sequence or escalation path: the behavior is goal-effective, yet the trajectory violates the safety boundary precisely because competence and constraint satisfaction come apart. For temporal $\times$ regulatory, a concrete failure mode is a system whose release-time evidence package remains formally on file even after tool semantics, retrieved knowledge, or escalation workflows have changed, so the deployment still appears compliant while the evidential basis for that claim has already expired. Relative to existing surveys, our taxonomy differs by organizing the field around these validation objects rather than around capabilities or application domains. \citet{luo2025} and \citet{yehudai2025} foreground evaluation gaps but do not make temporal validity or regulatory legibility first-class dimensions; \citet{zhao2026} captures safety and governance pressures in healthcare but does not generalize them as a cross-domain taxonomy.

\section{Survey of Existing Approaches}
\label{sec:approaches}

This section maps concrete validation approaches onto the five taxonomy dimensions and identifies \add{dense and sparse pairings} in the coded corpus (RQ3). Unlike Section~\ref{sec:related}, which positions adjacent literatures, the emphasis here is evidentiary: which approaches cover which validation objects\add{ among the coded papers}, and where \add{asymmetries appear}.

\subsection{Behavioral Validation}

\add{Behavioral validation is the most heavily represented dimension in the coded corpus (105 of 262 primary papers).} Benchmarks such as SWE-bench and ITBench, LLM-as-judge methods, and newer multi-agent suites have made capability measurement substantially richer, but the dominant target remains task success or broad process scoring rather than whether trajectories remain acceptable, consistent, and stable across semantically equivalent cases. SWE-bench and ITBench establish software-engineering task performance~\citep{swerebench2025,itbench2025}, while LLM-as-judge and process-scoring methods broaden the measurement toolkit~\citep{llmjudge2025,costar2025}. Assessment frameworks offer further general evaluation designs~\citep{assessmentFramework2025,unifiedEvaluation2025}. Multi-agent and interactive suites extend coverage to coordination and environments~\citep{multiagentbench2025,mast2025}, including AgentVerse and Tau-bench~\citep{agentverse2023,taubench2024}. Benchmark-aging work qualifies the durability of those scores~\citep{liang2022helm}. Across domains, the pattern is the same: behavioral evaluation is broad in coverage but shallow in what it guarantees, because outcome scores scale more readily than trajectory auditing.

This limitation is visible even in the strongest contemporary benchmark families. AgentBoard, TheAgentCompany, and AgentHarm extend behavioral evaluation toward multi-turn interaction, consequential workflows, and harmful-action measurement, yet they still stop short of specifying lifecycle criteria for whether trajectories remain acceptable under change, handoff, or tool failure. AgentBoard and TheAgentCompany expose the interaction and workflow dimensions~\citep{agentboard2024,theagentcompany2025}, while AgentHarm measures harmful-action exposure~\citep{agentharm2024}. The same pattern recurs across domains: in clinical orchestration~\citep{autonomousAgentic2025}, software engineering benchmarks and analyses~\citep{evaluatingSoftware2025,bib:beyondchatbox}, and plan-execute-generate-judge pipelines~\citep{planexecutegeneratejudgeSelf2025}, cybersecurity~\citep{agenticSecurity2025,benchmarkingOffensive2025}, and finance and logistics~\citep{ablLlmbased2025,smartopsDemand2025}, outcome scores do not expose intermediate-step inconsistencies, tool-use brittleness, or non-deterministic plan selection. Broader evaluations of agent-based and generative deployments report the same asymmetry~\citep{evaluatingAgentbased2025,overviewGenai2025}, including autonomous-agent deployments~\citep{autonomousAgents2025}, and failure-analysis studies locate the breakdowns specifically in reasoning transitions and tool-invocation sequences rather than in final answers~\citep{exploringAutonomous2025,confidentBut2025}. Dynamic suites such as RefuteBench~2.0 and AutoONBench probe consistency under perturbation~\citep{refutebenchAgentic2025,autoonbenchBenchmark2025}. Agent-drift studies track degradation over repeated interaction~\citep{agentDrift2025,limitsVerification2016}, including in self-evolving settings~\citep{agenticSelfevolving2025}; trajectory-level auditing is nevertheless not a standard behavioral-validation requirement.

\subsection{Safety Validation}
\label{sec:approaches-safety}

Safety validation is active but fragmented. As Section~\ref{sec:taxonomy} noted, the Safety column aggregates functional-safety and adversarial-security work; in the approaches layer, most recent density comes from prompt injection, misuse containment, and runtime enforcement rather than explicit safe-stop or graceful-degradation design.\add{ Functional-safety papers appear mainly under benchmarks (13) and runtime monitors (12); CPS-validation and assurance-case appearances are almost entirely functional-safety, and no adversarial-security-only paper carries either tag (Table~\ref{tab:gap-matrix}).} RTBAS and G-Safeguard exemplify runtime enforcement~\citep{rtbas2025,gsafeguard2025}, whereas AgentGuard and foundational safety work motivate bounded
autonomy~\citep{agentguard2025,amodei2016concrete}. Partial progress comes from tool-use containment through access control, semantic gating, defense-in-depth architectures, and uncertainty-driven monitoring. MCP security and semantic gating constrain tool use~\citep{mcpsecureRuntime2025,adaptiveSemantic2025}; defense-in-depth and uncertainty monitoring add complementary controls~\citep{multilayeredSecurity2025,uncertaintydrivenMonitoring2025}. \add{Further progress comes from verifiable safeguards and safety-requirement derivation.} Explainable and unified-safeguard approaches make local controls inspectable~\citep{explainableVerifiable2025,agentsafeUnified2025}; security and assurance frameworks translate them into broader claims~\citep{securityFramework2025,safetyAssurance2025}. Safety-requirement derivation and calibration work address the specification side~\citep{derivingSafety2025,kadavath2022knowing}, with debate-based approaches providing a related mechanism~\citep{debate2025}. Multimodal and multi-agent defenses broaden the threat coverage~\citep{multimodalFramework2025,unifiedFramework2025}; runtime resonance and moderation systems add operational controls~\citep{resonateRuntime2025,agenticModeration2025}, alongside cybersecurity-focused defenses~\citep{bib:cybersecurity}.

The remaining gap is not intervention in general, but a unified specification for safe-stop, graceful degradation, and reviewable evidence generation across diverse agent architectures and toolchains. Even systems such as RTBAS, G-Safeguard, and AgentGuard treat agents as intervenable information-flow systems: RTBAS and G-Safeguard provide runtime controls~\citep{rtbas2025,gsafeguard2025}, and AgentGuard extends them to agentic execution~\citep{agentguard2025}. They still leave open how deployments should hand off, roll back, or degrade when confidence collapses, or safeguards disagree.

\subsection{Temporal and Lifecycle Validation}

\add{Temporal validation is the least represented dimension in the coded corpus (14 primary papers, 5.3\%), and its 20 appearances are spread over five approach families, most often assurance cases (8) and MLOps/drift practices (5).} Benchmark-contamination work shows that static suites age, while MLOps, CPS, and regulatory literatures show that deployed systems require evidence maintenance, change control, and post-deployment monitoring rather than one-time benchmark results. Benchmark contamination and MLOps work show why one-time scores become stale~\citep{benchcontam2025,mlops2023}; expert monitoring and dataset-shift research identify the signals that should prompt renewal~\citep{expertmonitoring2024,quionero2009dataset}. Hidden-debt and production-ML studies explain the operational sources of this decay~\citep{sculley2015hidden,breck2017ml}, while \add{FDA guidance on predetermined change control and EU post-market surveillance guidance make change control and post-market evidence explicit expectations for AI-enabled medical devices}~\citep{fda-pccp2025,mdcg2025-10}. What is still missing is operationalization for agentic systems: what to monitor, when to revalidate, and how to refresh evidence when memory, tools, or workflows change.

Recent work makes this gap concrete without closing it. Agent-drift and long-horizon studies show degradation over repeated interaction or after many locally acceptable steps~\citep{agentDrift2025,memoryt1Reinforcement2025}, including explicitly temporal agent settings~\citep{agenticTemporal2025}. MLOps monitoring detects performance change, feature drift, and retraining triggers, but usually assumes stable prediction targets rather than open-ended policies that evolve through tool use, interaction, and memory adaptation. Observability and monitoring-maintenance frameworks provide the core instruments~\citep{observabilityFramework2025,maintainingMonitoring2025}; anomaly and reliability studies address their operational use~\citep{yourAnomaly2025,improvingReliability2025}, alongside model-updating methods~\citep{bib:updatingml}. Agent-specific harnesses and protocols increasingly treat lifecycle change as expected~\citep{ahe2026,autogenesis2026}, \add{and healthcare-agent reviews call for post-deployment drift monitoring~\citep{collaco2026,sharma2026}, a capability that a review of 49 healthcare agents finds almost entirely absent~\citep{bib:seventax}, and digital-twin work proposes continuous virtual validation for autonomous driving~\citep{veledar2019dt}.} The core temporal gap is therefore the missing bridge between observability and obligation: when a tool-schema revision invalidates earlier evidence, when memory contamination requires a fresh evaluation window, or when monitored drift becomes assurance failure.

\subsection{Regulatory and Assurance Approaches}

The regulatory dimension is addressed primarily by governance and assurance-case frameworks rather than technical methods. \add{In the US, final FDA guidance on predetermined change control plans makes change control an explicit expectation for AI-enabled devices~\citep{fda-pccp2025}, and the non-binding draft lifecycle guidance extends this expectation to lifecycle evidence~\citep{fda-aiedsflm2025}; in the EU, the MDR imposes binding post-market obligations~\citep{eu-mdr} that non-binding MDCG guidance explains, including their interplay with the AI Act~\citep{mdcg2025-6,mdcg2025-10}.} IEEE guidance and assurance-case methods provide structured argumentation~\citep{ieeeguide2025,alelyani2025}, but the engineering implementation remains underspecified for agentic systems. Put concretely, the surveyed governance documents\add{, binding or not,} are comparatively clear about the evidentiary \add{expectations}, such as traceability, change control, post-market evidence, and reviewable justification. However, they are far less specific about how an agentic system should operationally package trajectory evidence, monitoring outputs, override records, and revalidation triggers into a reviewable evidential payload. Mature argument patterns already exist across regulated domains, autonomous inspection, unmanned aircraft, spacecraft, automotive systems, and safety-critical machine learning. Reference patterns and interlocking-safety work establish reusable assurance structures~\citep{patternsReference2025,interlockingSafety2025}; software and formal assurance-case studies carry them into regulated engineering~\citep{softwareAssurance2025,assuranceCase20252}. Automotive and security applications demonstrate domain transfer~\citep{applicationAutomotive2025,quantitativeSecurity2025}, including for safety-critical machine learning~\citep{assuranceAimlbased2025}. A second body of work makes the assurance stack more operational through automated case management and continuous assurance~\citep{automatedModelbased2025,continuousDeployment2025}, model-risk governance and policy pipelines~\citep{modelRisk2025,policyPipeline2025}, and automated interpretation~\citep{automatedInterpretation2025}. Governance-oriented analyses extend the picture to liability and decentralized accountability~\citep{liabilityProblem2025,decentralizedGovernance2025}, and agentic-specific assurance proposals begin to translate these obligations toward agent deployments. Reliability-by-design and safety-landscape studies make this move explicitly~\citep{reliabilitybydesignAgentic2025,landscapeSafety2025}; agentic assurance cases and holistic frameworks extend it~\citep{assuranceCase2025,holisticAssurance2025}, alongside broader assurance-case work~\citep{bib:assurancecase}. All of these still stop short of specifying the trajectory-evidence bundle for open-ended agentic systems.

\subsection{Multi-Agent Validation}

Multi-agent validation is \add{concentrated in benchmark-style evidence}.\add{ In the coded corpus it has 47 primary papers (17.9\%) and 49 appearances, of which 36 are benchmark-type; runtime-monitor (10) and CPS-validation (3) appearances are few, and assurance-case and MLOps/drift appearances are absent.} Benchmarks, failure-mode studies, and runtime monitoring now provide useful data on interaction quality, yet they do not yield compositional validation certificates for open-ended LLM coordination. MultiAgentBench and MAST measure coordination outcomes and failure modes~\citep{multiagentbench2025,mast2025}, while G-Safeguard supplies a runtime perspective~\citep{gsafeguard2025}. Formal verification and distributed-systems work provides candidate compositional methods~\citep{verifyingMulti2025,formalizingDistributed2025}, including engineering and multilevel protocol approaches~\citep{engineeringMultiagent2025,multilevelProtocol2025}. Classical coordination frameworks offer additional precedents~\citep{camel2023,bib:auctionmas}, with recent design work extending them~\citep{dawnDesigning2025}. Recent evidence from asynchronous software-engineering agents makes the same point in a practically important setting: coordination quality depends on delegation structure, synchronization discipline, and handoff management, not only on single-agent competence~\citep{caid2026}. Classical multi-agent systems (MAS) verification remains methodologically relevant but only partially transferable because LLM-based agents operate over open-ended natural language and learned policies rather than bounded programs and finite vocabularies. The central gap is therefore structural: the challenge is not merely to test more agents, but to certify interaction-level properties that emerge only at the system level.

Supporting evidence converges from three further directions. Studies of collaborative planning, auction-based allocation, and decentralized coordination show that individually competent agents still produce oscillation, deadlock, or degraded joint behavior. LLM and reactive-coordination studies document these effects~\citep{canLlm2025,reactiveMultiagent2025}, as do behavior-tree approaches~\citep{behaviorTree2025}. Trust modeling and role-specialized traceability improve partner selection and post-hoc auditability without yielding compositional guarantees. Task-specific and contextual trust models improve partner selection~\citep{taskspecificTrust2025,contextualTrust2025}; traceability and comparative studies improve auditability~\citep{traceabilityAccountability2025,comparativeSurvey2025}, including in scientific scaling settings~\citep{scienceScaling2025}. And digital-twin simulation together with cascade-aware analysis \add{could supply} the substrate on which interaction-level hazards can be exercised before deployment~\citep{veledar2019dt,sambasivan2021cascades}. The common conclusion is that collective failures and interaction-level hazards cannot be inferred from single-agent scores alone.

\begingroup
\begin{table*}[!tbp]
\centering
\caption{Approach families by primary validation dimension: appearances (and \% of the column's papers). Safety is split by sub-strand (functional, adversarial, both). Papers may carry several tags, so percentages can exceed 100\%; the six families are not exhaustive. Counts reflect publication volume, not maturity.}
\label{tab:gap-matrix}
\footnotesize
\setlength{\tabcolsep}{3pt}
\renewcommand{\arraystretch}{1.08}
\begin{tabularx}{\textwidth}{@{}>{\raggedright\arraybackslash}Xccccccccc@{}}
\toprule
 & & \multicolumn{4}{c}{\textbf{Safety}} & & & & \\
\cmidrule(lr){3-6}
\textbf{Approach family} & \textbf{Behav.} & \textbf{Funct.} & \textbf{Advers.} & \textbf{Both} & \textbf{All} & \textbf{Temp.} & \textbf{Reg.} & \textbf{Multi-ag.} & \textbf{Total} \\
\midrule
Benchmarks & 97 (92) & 13 (57) & 21 (72) & 4 (44) & 38 (62) & 4 (29) & 8 (23) & 36 (77) & 183 \\
Runtime Monitors & 8 (8) & 12 (52) & 13 (45) & 7 (78) & 32 (52) & 2 (14) & 6 (17) & 10 (21) & 58 \\
CPS Validation & 9 (9) & 5 (22) & 0 (0) & 1 (11) & 6 (10) & 1 (7) & 3 (9) & 3 (6) & 22 \\
Assurance Cases & 0 (0) & 3 (13) & 0 (0) & 0 (0) & 3 (5) & 8 (57) & 20 (57) & 0 (0) & 31 \\
MLOps/Drift & 2 (2) & 0 (0) & 0 (0) & 2 (22) & 2 (3) & 5 (36) & 2 (6) & 0 (0) & 11 \\
Reg Guidance & 0 (0) & 0 (0) & 0 (0) & 0 (0) & 0 (0) & 0 (0) & 13 (37) & 0 (0) & 13 \\
\midrule
Appearances & 116 & 33 & 34 & 14 & 81 & 20 & 52 & 49 & 318 \\
Papers & 105 & 23 & 29 & 9 & 61 & 14 & 35 & 47 & 262 \\
Unmatched papers & 3 & 0 & 2 & 1 & 3 & 0 & 4 & 5 & 15 \\
\bottomrule
\end{tabularx}
\end{table*}
\endgroup

\begin{table}[!htbp]
\centering
\caption{Relation between the two counting layers: paper-level primary-dimension counts (single label, total \add{262}) and gap-matrix appearance counts (multi-label over approach families, \add{}total \add{318}).}
\label{tab:count-reconciliation}
\footnotesize
\setlength{\tabcolsep}{3pt}
\renewcommand{\arraystretch}{1.08}
\begin{tabularx}{\columnwidth}{@{}>{\raggedright\arraybackslash}p{0.23\columnwidth}>{\raggedright\arraybackslash}p{0.22\columnwidth}>{\centering\arraybackslash}p{0.08\columnwidth}X@{}}
\toprule
\textbf{Counting layer} & \textbf{Behavioral / Safety / Temporal / Regulatory / Multi-agent} & \textbf{Total} & \textbf{What is being counted} \\
\midrule
Primary dimension (paper-level; single label) & \add{105 / 61 / 14 / 35 / 47} & \add{262} & Each included paper carries exactly one primary dimension under the codebook; these counts ground all paper-level claims, the year-band composition in Figure~\ref{fig:year-dist}, and the sensitivity bounds in Table~\ref{tab:sensitivity-bounds}. Reported in Section~\ref{sec:corpus-composition}. \\
\midrule
Gap-matrix coverage (\add{approach-family appearances placed in the primary-dimension column; multi-label over families only}) & \add{116 / 81 / 20 / 52 / 49} & \add{318} & Appearance counts used for the coverage matrix in Table~\ref{tab:gap-matrix}: a paper may contribute to several approach-family cells\add{, but all of its appearances are placed in the column of its primary dimension; secondary dimensions are recorded but not counted}. The six approach families are not exhaustive, so papers whose approach fits none of them contribute no appearances. The matrix therefore counts \add{family-tag appearances}, not unique-paper frequency; \add{the sensitivity bounds in Table~\ref{tab:sensitivity-bounds} apply to the paper-level counts, whereas Table~\ref{tab:matrix-labelings} shows how this layer changes under alternative primary labels}. \\
\bottomrule
\end{tabularx}
\end{table}

\begingroup
Table~\ref{tab:count-reconciliation} relates the two counting layers: one primary dimension per paper versus multi-label approach-family appearances, which measure how often a kind of evidence appears rather than how many papers address a dimension. Read together with Table~\ref{tab:gap-matrix}, \add{the two layers agree on the largest and smallest dimensions but not on the middle of the ordering: by papers, multi-agent work (47) exceeds regulatory work (35), whereas by appearances regulatory work (52) exceeds multi-agent work (49), because regulatory papers more often carry several family tags}. Behavioral work leads (105 papers, 116 appearances, 97 of them benchmarks); safety follows (61 papers, 81 appearances, concentrated in benchmarks and runtime monitors); regulatory work (35 papers, 52 appearances) is dominated by assurance cases (20) and regulatory-guidance analysis (13); multi-agent work (47 papers, 49 appearances) rests mostly on benchmarks (36); and temporal work is the least represented on both layers (14 papers, 20 appearances spread over five families). Benchmark-style behavioral evaluation is thus the dominant form of evidence in the coded corpus, \add{regulatory evidence consists mainly of assurance cases and regulatory analysis, and multi-agent evidence is mainly benchmark-based, with few runtime-monitor and no assurance-case or MLOps/drift appearances among the coded papers}.
\endgroup

\begingroup
\label{q:cautions}Four cautions qualify this reading. First, percentages are taken over the papers of the column's primary dimension (Table~\ref{tab:gap-matrix}). Second, 15 of 262 papers (5.7\%) match none of the six families: multi-agent 5 of 47, regulatory 4 of 35, safety 3 of 61, behavioral 3 of 105, and temporal 0 of 14. The thin temporal column is therefore not an artifact of temporal papers falling outside the scheme, but the families were fixed a priori and are not exhaustive, so evidence types such as formal verification of agent protocols are not counted. Third, several cells depend on IEEE-only papers (Section~\ref{sec:methodology-limits}). Fourth, part of the matrix structure follows from the coding scheme rather than from the literature. The behavioral dimension (capability, benchmark, or trajectory-success measurement) overlaps with the Benchmarks family, and the safety dimension (runtime constraint enforcement, containment) with the Runtime Monitors family, so the 97 benchmark appearances among behavioral papers and the 32 runtime-monitor appearances among safety papers are partly expected by construction. Likewise, papers centered on regulatory obligations or on an assurance case are coded as regulatory, so all 13 Reg Guidance appearances and 20 of the 31 assurance-case appearances fall in that column, and the Behavioral $\times$ Assurance Cases cell is empty partly for this reason. Such cells describe how the coding scheme partitions the corpus. More generally, because a paper's appearances are counted only in its primary-dimension column, a sparse cell indicates few coded papers of that kind rather than the absence of a capability in the field.
\endgroup

\section{Illustrative Case Studies Across Operational Domains\add{: Constructed Traces}}
\label{sec:casestudy}

The five-dimension taxonomy is domain-general, but its force is clearest when applied to deployment settings where agentic systems already act consequentially. We present three parallel case studies, namely medical care delivery, industrial operations, and smart-mobility CPS, as matched illustrations of the same validation pressures. Each surfaces heterogeneous tool use, temporally evolving evidence, safety-critical intervention, human override, and operational accountability. \add{Prior clinical, healthcare-agent, and closed-loop CPS studies document related risks and failure mechanisms in these settings~\citep{collaco2026,zhao2026,sharma2026,draelos2026,lin2025iccps,brown2025iccps}, and the case studies below show how the taxonomy applies to them.}

Each case study is grounded in the reviewed literature and pairs the dimension mapping with a constructed trajectory trace: a step-by-step walkthrough using the per-step $(s_t,a_t,m_t)$ notation from the trajectory \add{$\tau$ of Eq.~\eqref{eq:trajectory}} defined in Section~\ref{sec:definitions}, consistent with the \add{failure mechanisms documented in the cited studies}. The constructed traces show where each latent defect would be intercepted and by which monitor\add{, making the taxonomy concrete at step granularity; they combine documented mechanisms with defects and thresholds that we introduce, and Section~\ref{sec:evidence-status} gives the evidence status of each claim.} \add{The traces illustrate how} components can pass their local tests while the trajectory as a whole fails.

\subsection{Evidence Status of the Case-Study Claims}
\label{sec:evidence-status}

\begingroup
Table~\ref{tab:evidence-status} classifies each claim used in the case studies. \emph{Observed} means reported empirically in the cited work; \emph{literature-supported} means a mechanism described in prior work that we reuse; \emph{constructed} means a combination or defect we introduce for illustration; \emph{illustrative} means a policy threshold chosen as an example. The support column shows how each cited work informs the case-study claim.

\begin{table*}[!tbp]
\centering
\caption{Evidence status of claims used in the constructed case studies. Only rows marked ``Observed'' are empirical findings of the cited work, at the level stated.}
\label{tab:evidence-status}
\footnotesize
\setlength{\tabcolsep}{3pt}
\renewcommand{\arraystretch}{1.1}
\begin{tabularx}{\textwidth}{@{}>{\raggedright\arraybackslash}p{1.5cm}>{\raggedright\arraybackslash}X>{\raggedright\arraybackslash}p{2.6cm}>{\raggedright\arraybackslash}p{3.4cm}@{}}
\toprule
\textbf{Case} & \textbf{Claim} & \textbf{Status} & \textbf{Support} \\
\midrule
Medical & LLMs can give unsafe or insufficiently escalatory answers to patient-posed medical questions & Observed (response level, not agentic trajectories; insufficient escalation shown by examples, not rated separately) & \citep{draelos2026} \\
\midrule
Medical & Healthcare agents are embedded in clinical information systems and socio-technical workflows & Literature-supported & \citep{zhao2026} \\
\midrule
Medical & Advice is drafted from a renal panel 105 minutes old & Constructed & Motivated by workflow coupling; not reported in \citep{draelos2026} \\
\midrule
Medical & Equivalent phrasings yield different triage classifications & Constructed; mechanism literature-supported & Prompt sensitivity of LLM responses~\citep{draelos2026} \\
\midrule
Industrial & Specialized agents coordinate commissioning and troubleshooting of drives & Literature-supported & \citep{driveaiagentMultiagent2025} \\
\midrule
Industrial & Concept drift degrades learned models, and labels for detecting it are often scarce in industrial settings & Literature-supported & \citep{connectionBetween2025,boundaryawareConcept2025} \\
\midrule
Industrial & Cached snapshot predates a firmware update; unsafe recommendations reach UAR 0.08; duplicated restart issued & Constructed combination & Composition of the mechanisms above \\
\midrule
Mobility & A highway agent consumes heterogeneous road and sensor signals & Literature-supported & \citep{multimodalAgentic2025} \\
\midrule
Mobility & OOD detection is widely adopted for learning-enabled CPS but does not by itself imply a safety violation; runtime safety monitoring and quantitative runtime verification have been demonstrated in simulation and on a physical testbed & Literature-supported & \citep{lin2025iccps,brown2025iccps} \\
\midrule
Mobility & Fog-onset episode with stale map layer, missed handoff budget, non-replayable inputs & Constructed combination & Composition of the mechanisms above \\
\midrule
All & Freshness limit 60 min, UAR threshold 0.02, CSR threshold 0.95 & Illustrative policy values & None; example values only \\
\bottomrule
\end{tabularx}
\end{table*}
\endgroup

\subsection{Metric Definitions for the Case Studies}\label{sec:metrics}

Three metrics appear throughout the illustrative case studies and policy vignettes. \add{Their operational definitions are ours, derived from the metrics listed in the taxonomy (Table~\ref{tab:taxonomy}); they are given here} to clarify the numerical statements in the traces that follow.

\begin{table}[!htbp]
\caption{Operational definitions of three illustrative metrics used in case-study traces. These definitions are derived from the taxonomy discussion in Section~\ref{sec:taxonomy}; the thresholds shown in the case studies (Section~\ref{sec:casestudy}) are example policy values, not empirical benchmark results.}
\label{tab:metrics-definitions}
\footnotesize
\setlength{\tabcolsep}{2.5pt}
\renewcommand{\arraystretch}{1.15}
\begin{tabularx}{\columnwidth}{@{}
>{\raggedright\arraybackslash}p{1.8cm}
>{\hsize=1.3\hsize\raggedright\arraybackslash}X
>{\hsize=0.7\hsize\raggedright\arraybackslash}X
@{}}
\toprule
\textbf{Metric} & \textbf{Definition} & \textbf{Evaluation window and interpretation} \\
\midrule
Evidence Freshness Interval (EFI) & Time elapsed since the last refresh of a data source (e.g., laboratory panel, sensor reading, or retrieved evidence) used in a trajectory decision. Measured in minutes or hours from the source's timestamp to the decision point. & Evaluation window: from source creation to decision time. Interpretation: if EFI exceeds a policy-specified freshness limit (e.g., 60 minutes for medication-relevant labs), the decision is flagged or blocked until the source is refreshed. Example threshold: EFI $= 105\,\text{min} > 60\,\text{min policy limit}$ $\rightarrow$ \textit{block}. \\
\midrule
Unsafe-Action Rate (UAR) & Proportion of executed \add{or recommended }actions (typically tool invocations or control commands\add{, or actuator-setting recommendations}) within a session or trajectory that violate safe-operating constraints or authorization policy. Numerator: count of unsafe actions. Denominator: total actions executed\add{ or recommended} in the window\add{, with the action type fixed per deployment}. & Evaluation window: rolling window over the current session or trajectory (typically 10–100 steps). Interpretation: if UAR exceeds a threshold (e.g., 0.02 for safety-critical systems), \add{escalation and revalidation are triggered; an individual action that violates a hard constraint is blocked directly, independently of the rate}. Example threshold: UAR $= 0.08 > 0.02\,\text{plant threshold}$ $\rightarrow$ \add{\textit{escalate}}. \\
\midrule
Coordination Success Rate (CSR) & Proportion of inter-agent or human–agent handoff events that complete without conflict, deadlock, or timeout. Numerator: count of conflict-free transitions. Denominator: total handoff events attempted in the evaluation window. & Evaluation window: rolling count of handoffs in the current session or deployment period. Interpretation: if CSR falls below a threshold (e.g., 0.95 for safety-critical coordination), escalation is triggered to reduce the coordination load or involve human oversight. Example threshold: CSR $= 0.91 < 0.95$ (deployment threshold\add{; 10 of 11 handoffs in the session}) $\rightarrow$ \textit{flag}. \\
\bottomrule
\end{tabularx}
\end{table}

\noindent\textbf{Note on illustrative thresholds.} The numerical examples in the metric definitions above (60 minutes, 0.02, 0.95) are representative policy-threshold values used only in the illustrative case-study traces. They are \emph{not} empirical results from agentic-system benchmarks and should not be interpreted as recommendations for real deployments. Each operational context requires calibration of these thresholds based on risk tolerance, domain constraints, and deployment-specific data.

\begingroup
\label{q:construct}Even as illustrations, the three metrics are operational proxies with known limits. Freshness does not guarantee validity: a recent source can be wrong and an old one can remain valid (e.g., static demographics), so EFI limits are claim-specific. UAR needs labeled unsafe actions, which requires stating the constraint set used as reference (authorization policy, operating envelope, or clinical rule), who labels, and inter-labeler agreement. An unweighted rate can hide rare but severe events, so UAR should be reported together with a severity-weighted variant and the count of severity-critical events, for which a zero-tolerance rule is more appropriate than a rate. Thresholds must be calibrated from hazard analysis and risk acceptance and checked on replayed or simulated traces for their false-block and miss rates; the values here are not calibrated. Finally, two questions must be kept apart. The first is a \emph{decision}: whether to block the current action. It should follow hazard analysis, so a single action that violates a hard constraint is blocked at once, whatever the rate and without waiting for a minimum count. The second is an \emph{inference}: what long-run unsafe-action rate the session reveals. The industrial trace's $0.08$ is 2 of 25 recommendations; a 95\% Wilson interval of roughly $0.02$--$0.25$ would apply only if the 25 recommendations were independent trials with a common probability, an assumption that sequential actions within one session and overlapping rolling windows usually violate, so the interval is optimistic. A rate threshold such as $0.02$ is therefore best used for trend-based escalation and revalidation, not as the sole trigger for blocking.
\endgroup

\subsection{Medical Care Delivery}

A representative deployment target \add{in the literature} is postoperative nursing support with LLM-agent systems, where agents assist observation, follow-up, and coordination around postoperative care workflows~\citep{parallelNursing2025}\add{, illustrated there on synthetic patient cases}. The most directly documented failure evidence comes from Draelos et al., who show that large language models can provide unsafe answers to patient-posed medical questions, including advice that is clinically unsafe or insufficiently escalatory for the presenting condition~\citep{draelos2026}\add{; unsafe responses are rated as a category, whereas insufficient escalation is documented through qualitative examples}. While Draelos et al.\ examine LLM responses rather than full agentic trajectories, their findings document unsafe and insufficiently escalatory response patterns. In the constructed trace below, we treat those response-level failures as local breakdowns that an agentic clinical deployment could amplify through retrieval, escalation, and follow-up. The stale-evidence step is an illustrative trajectory-level risk motivated by clinical workflow coupling rather than a direct incident reported by Draelos et al. In the five-dimension structure, this failure is \emph{behavioral} because similar symptom descriptions can produce different advice trajectories; \emph{safety} because the recommendation itself can be unsafe; \emph{temporal} because the correctness of the advice depends on whether medication lists, recent vitals, or symptom progression are still current; \emph{regulatory} because any recommendation that affects care requires reviewable justification; and \emph{multi-agent} because the effective outcome depends on coordination across the agent, electronic health record (EHR) tools, nursing staff, physicians, and pharmacy. \add{Zhao et al.\ likewise describe healthcare agents as embedded in clinical information systems (EHR, imaging, and laboratory systems) and frame their adoption in socio-technical terms.} \add{A healthcare-agent review and expert-monitoring work from MLOps motivate these indicators~\citep{zhao2026,expertmonitoring2024}. A capability taxonomy of 49 healthcare agents finds drift detection and mitigation almost always absent~\citep{bib:seventax}; an agentic diabetes framework illustrates the step from prediction to action, although its evaluation covers predictor and simulator accuracy and use-case demonstrations~\citep{knowledgeAction2025}; and a home-based autism screening system adds an LLM layer that turns classifier output into caregiver feedback~\citep{empoweringConvenient2025}.} The validation lesson is that clinical usefulness cannot be separated from freshness, escalation timing, and documented responsibility for action.

Table~\ref{tab:trace-medical} makes the mapping operational through a constructed postoperative-support trajectory. Every component in this trace can pass its local tests; the retrieval API returns well-formed records; the classifier meets its accuracy target on its test distribution; the generated message is fluent, while the trajectory as a whole is unsafe. Each latent defect is intercepted, if at all, by a different dimension, and the first interception (the freshness breach at $t_1$) precedes the clinically dangerous step at $t_3$: trajectory-level validation matters precisely because it can stop a defective episode before its consequential action.

\begin{table}[!htbp]
\caption{Constructed trajectory trace for the medical case study, consistent with the response-level failure patterns in~\citep{draelos2026} and the socio-technical coupling in~\citep{zhao2026}. Each row is one step $(s_t,a_t,m_t)$; the last column names the taxonomy dimension and monitor signal that intercepts the defect. \add{Rows after a \emph{block} or \emph{escalate} show the unmitigated counterfactual trajectory, i.e., what would follow if the intervention were not enforced, to illustrate which further monitors would fire. }Thresholds are illustrative policy values, not benchmark results.}
\label{tab:trace-medical}
\footnotesize
\setlength{\extrarowheight}{3pt}
\begin{tabularx}{\columnwidth}{@{}c >{\raggedright\arraybackslash}X >{\raggedright\arraybackslash}p{2.6cm} >{\raggedright\arraybackslash}p{3.6cm}@{}}
\toprule
$t$ & Trajectory event $(s_t,a_t,m_t)$ & Latent defect & Dimension $\rightarrow$ signal and outcome \\
\midrule
\add{$t_{0}$} & Patient message reports a new post-operative symptom; the agent opens a session and initializes state $s_0$. & --- & --- \\
\midrule
\add{$t_{1}$} & Agent retrieves chart and laboratory data via the EHR tool; $m_1$ records that the renal panel timestamp is 105 minutes old. & Freshness policy for medication-relevant decisions is 60 minutes. & Temporal $\rightarrow$ EFI $=105>60$ min: \emph{block} pending refresh. \\
\midrule
\add{$t_{2}$} & Planner classifies the report as routine discomfort; a replay run with a semantically equivalent phrasing yields a different classification. & Trajectory inconsistency under paraphrase. & Behavioral $\rightarrow$ consistency check on replayed trajectories: \emph{flag}. \\
\midrule
\add{$t_{3}$} & Agent drafts medication-adjustment advice and does not escalate to clinical staff. & Advice insufficiently escalatory for the presenting differential, the pattern documented in~\citep{draelos2026}. & Safety $\rightarrow$ escalation-appropriateness rule: \emph{escalate}. \\
\midrule
\add{$t_{4}$} & Memory update logs the episode as resolved without linking the advice to the chart snapshot used. & Decision rationale not reconstructable. & Regulatory $\rightarrow$ trace-completeness audit: \emph{flag}. \\
\midrule
\add{$t_{5}$} & Follow-up task enters the nursing queue with low priority and misses the shift handover. & Mistimed coordination between agent and care workflow. & Multi-agent $\rightarrow$ handoff-latency monitor: \emph{escalate}. \\
\bottomrule
\end{tabularx}
\end{table}

\subsection{Industrial Operations}

A representative industrial target is multi-agent support for commissioning and troubleshooting industrial drives, where specialized agents coordinate diagnosis steps, parameter checks, and operator-facing recommendations during equipment setup and fault analysis~\citep{driveaiagentMultiagent2025}.\add{ That system is evaluated on synthetic scenarios against expert-defined configurations and manual entries, and its safety risks are left to future work; the dimension reading that follows is ours.} The five-dimensional structure recurs end to end: \emph{behavioral}, because similar fault signatures can trigger different diagnostic decompositions or recovery plans; \emph{safety}, because an incorrect recommendation can propagate into unsafe actuator settings or ill-timed restart advice; \emph{temporal}, because telemetry, maintenance history, and machine state can drift while the troubleshooting loop is still running; \emph{regulatory}, because industrial interventions require a reconstructable rationale for why a suggested action was issued; and \emph{multi-agent}, because the outcome depends on coordination across specialized software agents, plant-control tools, operators, and the physical asset itself. \add{Industrial concept-drift work supports one part of this mapping: drift degrades learned models after deployment, and the labels needed to detect it are often scarce~\citep{connectionBetween2025,boundaryawareConcept2025}. Neither study concerns agents, and both evaluate on synthetic or public benchmark streams. Healthcare-agent reviews likewise recommend post-deployment drift monitoring~\citep{collaco2026,sharma2026}, which suggests that the requirement is not specific to industrial settings.} Here the important shift is from evaluating whether a recommendation looks plausible to evaluating whether it remains safe and accountable while plant state, operators, and controllers continue to change.

Table~\ref{tab:trace-industrial} traces a constructed commissioning episode. The trace makes the drift literature concrete: the defect introduced at $t_1$ is temporal, but its consequential expression at $t_3$ is a safety violation, and the only dimension that explains the duplicated restart at $t_4$ is multi-agent coordination. A validation regime checking any single dimension would have caught one defect and missed the episode.

\begin{table}[!htbp]
\caption{Constructed trajectory trace for the industrial case study, consistent with the multi-agent drive-commissioning setting of~\citep{driveaiagentMultiagent2025} and the industrial drift and uncertainty findings of~\citep{connectionBetween2025,boundaryawareConcept2025}. \add{Rows after a \emph{block} show the unmitigated counterfactual trajectory, i.e., what would follow if the block were not enforced. }Thresholds are illustrative policy values.}
\label{tab:trace-industrial}
\footnotesize
\setlength{\extrarowheight}{3pt}
\begin{tabularx}{\columnwidth}{@{}c >{\raggedright\arraybackslash}X >{\raggedright\arraybackslash}p{2.6cm} >{\raggedright\arraybackslash}p{3.6cm}@{}}
\toprule
$t$ & Trajectory event $(s_t,a_t,m_t)$ & Latent defect & Dimension $\rightarrow$ signal and outcome \\
\midrule
\add{$t_{0}$} & An intermittent overcurrent fault is reported during drive commissioning; the diagnostic session opens. & --- & --- \\
\midrule
\add{$t_{1}$} & Telemetry tool returns a cached parameter snapshot predating a recent firmware update; $m_1$ records the snapshot version. & Tool output reflects a superseded machine configuration. & Temporal $\rightarrow$ evidence-version check: \emph{flag}. \\
\midrule
\add{$t_{2}$} & Planner decomposes the diagnosis; a replay of the same fault signature yields a different tool-invocation order\add{, in which the parameter-change recommendation precedes the telemetry-version check}. & \add{Requirement R1 (the telemetry version is verified before any parameter-change recommendation) is violated in one replayed ordering; orderings satisfying R1 are acceptable.} & Behavioral $\rightarrow$ \add{requirement-conformance check of R1 on replayed trajectories}: \emph{flag}. \\
\midrule
\add{$t_{3}$} & Agent recommends raising the current limit and issuing a restart; the session's unsafe-action rate over actuator-setting recommendations reaches 0.08\add{ (2 of 25 recommendations in the current 25-step window)}. & Recommendation violates the post-update safe operating envelope. & Safety $\rightarrow$ \add{safe-operating-envelope check: \emph{block}; }UAR $=0.08>0.02$ plant threshold: \add{\emph{escalate}}. \\
\midrule
\add{$t_{4}$} & The thermal-monitoring peer agent is not consulted; a duplicated restart instruction is issued through a second channel. & Conflicting inter-agent instructions to the same asset. & Multi-agent $\rightarrow$ duplicated-action detector: \emph{flag}. \\
\midrule
\add{$t_{5}$} & Operator accepts the remaining suggestion; the log does not link the recommendation to the telemetry version that produced it. & Audit cannot reconstruct why the action was issued. & Regulatory $\rightarrow$ provenance-completeness audit: \emph{flag}. \\
\bottomrule
\end{tabularx}
\end{table}

\subsection{Smart-Mobility CPS}

A representative mobility-side target is highway safety monitoring with multimodal agentic systems, where \add{a fixed pipeline of agents combines camera images, dashcam footage, vehicle sensors (airbag, seatbelt), and weather information to describe driving scenes and to generate accident reports}~\citep{multimodalAgentic2025}.\add{ That study rates scene descriptions and reports manually on small samples; using such a pipeline to flag hazards or request intervention, as in the trace below, is our extension.} The validation issues again map cleanly onto five dimensions: \emph{behavioral}, because small changes in perception inputs can alter downstream interpretations and recommended responses; \emph{safety}, because missed or spurious hazard classification directly affects roadway risk; \emph{temporal}, because the validity of a decision depends on whether traffic state, weather, and map context are still current; \emph{regulatory}, because public-safety actions need inspectable justification and replayable evidence; and \emph{multi-agent}, because the effective behavior is distributed across perception modules, runtime monitors, infrastructure services, and human supervisors. CPS work on \add{safety monitoring under out-of-distribution conditions} and perception-based quantitative runtime verification treats smart-mobility assurance as a runtime problem rather than a pre-deployment benchmark problem. ICCPS studies describe the closed-loop requirement~\citep{lin2025iccps,brown2025iccps}, and digital-twin work provides a complementary account~\citep{veledar2019dt}. This makes mobility the clearest reminder that locally sensible steps can still compose into globally unsafe behavior when timing, coordination, and intervention logic are wrong.

Table~\ref{tab:trace-mobility} traces a constructed fog-onset episode. It differs from the previous two traces in that the earliest signal ($t_1$) is already an escalation trigger: \add{OOD detection is widely adopted as a runtime instrument for learning-enabled CPS, and runtime safety monitoring and quantitative runtime verification have been demonstrated in simulation and on a physical F1Tenth testbed~\citep{lin2025iccps,brown2025iccps}. Because OOD inputs do not by themselves imply a safety violation~\citep{lin2025iccps}, the persistent OOD score, together with degraded perception confidence, triggers a request for review rather than asserting a hazard. Yet} the trace still accumulates four further defects that OOD monitoring alone does not see. \add{Coverage} in one dimension does not substitute for the others.

\begin{table}[!htbp]
\caption{Constructed trajectory trace for the smart-mobility case study, consistent with the multimodal highway-monitoring setting of~\citep{multimodalAgentic2025} and the runtime-assurance instruments of~\citep{lin2025iccps,brown2025iccps}. \add{Rows after an \emph{escalate} show the trajectory while the escalation is pending. }Thresholds are illustrative policy values.}
\label{tab:trace-mobility}
\footnotesize
\setlength{\extrarowheight}{3pt}
\begin{tabularx}{\columnwidth}{@{}c >{\raggedright\arraybackslash}X >{\raggedright\arraybackslash}p{2.6cm} >{\raggedright\arraybackslash}p{3.6cm}@{}}
\toprule
$t$ & Trajectory event $(s_t,a_t,m_t)$ & Latent defect & Dimension $\rightarrow$ signal and outcome \\
\midrule
\add{$t_{0}$} & Highway monitoring agent ingests camera and roadside-sensor feeds as fog develops over a monitored segment. & --- & --- \\
\midrule
\add{$t_{1}$} & Perception confidence degrades; the out-of-distribution score rises and persists; $m_1$ logs the monitor output. & Inputs are leaving the validated operating domain. & Safety $\rightarrow$ OOD-persistence monitor: \emph{escalate}. \\
\midrule
\add{$t_{2}$} & Agent classifies the obstruction as low-priority debris; a near-identical scene in replay is classified as a hazard. & Timing-dependent divergent interpretation of similar scenes. & Behavioral $\rightarrow$ consistency under scene perturbation: \emph{flag}. \\
\midrule
\add{$t_{3}$} & The deferral decision relies on a map layer older than its freshness bound. & Stale context informs a safety-relevant deferral. & Temporal $\rightarrow$ map-freshness check: \emph{flag}. \\
\midrule
\add{$t_{4}$} & Handoff to the infrastructure service misses its latency budget; the session's conflict-free handoff rate falls to 0.91\add{ (10 of 11 handoff events since the session opened)}. & Coordination below the deployment threshold. & Multi-agent $\rightarrow$ CSR $=0.91<0.95$: \emph{flag}. \\
\midrule
\add{$t_{5}$} & Escalation eventually occurs, but the inputs that drove the earlier deferral are not replayable for review. & Intervention record not legible to oversight. & Regulatory $\rightarrow$ audit-replay check: \emph{flag}. \\
\bottomrule
\end{tabularx}
\end{table}

\subsection{Cross-Domain Summary}

\newcolumntype{Y}{>{\raggedright\arraybackslash\footnotesize}X}

\begin{table}[!htbp]
\caption{Cross-domain summary of the five validation dimensions.}
\label{tab:case-cross-domain}
\footnotesize
\setlength{\extrarowheight}{4pt}
\begin{tabularx}{\columnwidth}{>{\bfseries\footnotesize\raggedright\arraybackslash}p{2.2cm} Y Y Y}
\toprule
\textbf{Dimension} & \textbf{Medical} & \textbf{Industrial} & \textbf{Smart Mobility} \\
\midrule
Behavioral &
Divergent tool-use and escalation trajectories for similar patient states &
Similar process states can trigger brittle planning or actuator sequencing &
Similar traffic scenes can produce unsafe trajectory choices under different timing assumptions \\
\midrule
Safety &
Unsafe recommendations, delayed holds, or missed escalation affect care &
Wrong sequencing or delayed intervention can damage equipment or violate limits &
Navigation, routing, or handoff errors create system-level risk \\
\midrule
Temporal &
Labs, device feeds, and orders age asynchronously &
Telemetry, controller state, and maintenance context drift during execution &
Sensor latency, map freshness, and traffic evolution alter safe decisions over time \\
\midrule
Regulatory &
Clinical interventions require traceable justification and reviewability &
Audits require change traceability, incident reconstruction, and operator accountability &
Public-safety actions require legible intervention records \\
\midrule
Multi-agent &
Outcomes depend on coordination among agent, EHR tools, clinicians, and pharmacy &
Agent, control tools, operators, and plant infrastructure form one coupled loop &
Safety emerges from coordination among agents, platforms, infrastructure, and supervisors \\
\bottomrule
\end{tabularx}
\smallskip

\raggedright\footnotesize
\textit{Illustrative policy vignettes.}
Medical: EFI $= 105$ min against a 60-min limit $\rightarrow$ \textit{block}.
Industrial: UAR $= 0.08$ against a threshold of $0.02$ $\rightarrow$ \add{\textit{escalate} (the unsafe recommendation itself is blocked by the safe-operating-envelope check)}.
Smart mobility: CSR $= 0.91$ against a threshold of $0.95$ $\rightarrow$ \textit{flag}.
These values are illustrative policy examples rather than reported benchmark results; they correspond to the threshold breaches at $t_1$, $t_3$, and $t_4$ of the worked traces in Tables~\ref{tab:trace-medical}--\ref{tab:trace-mobility}.
\end{table}

Across all three domains, the common pattern is that the agent is not an isolated predictor but a participant in a closed-loop socio-technical system. Failures emerge from stale evidence, unsafe intermediate actions, mistimed escalation, and coordination breakdowns, precisely why behavioral, safety, temporal, regulatory, and multi-agent validation must be addressed together rather than as separate checklists. The trajectory, not the final output, is the unit at which the relevant failure modes become visible. The worked traces (Tables~\ref{tab:trace-medical}--\ref{tab:trace-mobility}) make this concrete in a second way: the medical and industrial traces are both first intercepted by a temporal signal, yet only in the industrial trace does that stale-evidence defect \add{itself }go on to \add{cause} a downstream safety violation\add{ (the unsafe recommendation at $t_3$ follows from the superseded snapshot); in the medical trace the safety defect at $t_3$ is an escalation failure that would occur even with a fresh panel;} \add{the mobility trace, by contrast, is} intercepted first by runtime safety monitoring; no single instrument dominates across domains. The policy vignettes in Table~\ref{tab:case-cross-domain} show how the same thresholds convert dimension signals into actionable outcomes such as block, flag, or escalate.
What the case studies add beyond the taxonomy alone is \emph{cross-domain mapping} of the taxonomy onto three unlike deployment domains. \add{The cited literature documents related failure mechanisms in each domain, and the traces show how} the same five validation questions \add{arise} in medical care (\citealp{draelos2026,zhao2026}), industrial operations (\add{\citealp{driveaiagentMultiagent2025,connectionBetween2025}}), and smart mobility (\citealp{lin2025iccps,brown2025iccps}) despite their different hazards, tools, oversight regimes, and time scales. \add{This illustrates} \add{how the taxonomy applies to} deployment-level validation concerns across consequential domains.

\section{Research Agenda: Building the Validation Stack}
\label{sec:agenda}

The survey supports four cumulative directions for a validation stack: specification, stress exposure, runtime containment, and auditability.\add{ Directions 2--4 start from a sparse cell of the coverage matrix (Table~\ref{tab:gap-matrix}) and develop it into a research question; Direction 1 is motivated conceptually. All four directions are interpretive (Table~\ref{tab:evidence-layers}).}

\textbf{1) Bounded-autonomy specifications.} \emph{Which specification formalisms can express partial trajectory contracts for open-ended, tool-using agents without collapsing them into brittle scripts?} This direction \add{is motivated conceptually rather than by a matrix cell (the \emph{Behavioral $\times$ Assurance Cases} cell in Table~\ref{tab:gap-matrix} is empty largely by construction of the coding scheme; Section~\ref{sec:approaches})}. The gap is not measuring whether an agent can solve a task once; it is stating, in machine-checkable form, what the agent is allowed, forbidden, or obligated to do while pursuing that task.

The core research problem is to define \emph{partial} contracts over trajectories: obligations to escalate, regions of forbidden action, constraints on delegation, and memory-update rules that survive policy variation and environment change. In agentic systems, engineers rarely know the exact action sequence in advance; what they know instead is the structure of acceptable behavior under uncertainty. That makes the problem a requirements-language challenge at the boundary of temporal logic, assume–guarantee reasoning, runtime shielding, and socio-technical systems engineering rather than a mere benchmark-design exercise. Runtime shielding and safety-of-the-intended-functionality work supplies formal starting points~\citep{alshiekh2018safe,iso21448}; \add{healthcare-agent reviews show how thin the evidence from deployed agents still is~\citep{zhao2026,collaco2026},} while constitutional approaches offer a related agentic framing~\citep{bib:constitutional}.

The expected artifact is an \emph{interchange format} for bounded autonomy: a specification object that can be compiled into test oracles, runtime monitors, and assurance claims, evaluated through obligation coverage, violation-detection recall, and representational adequacy.

\begingroup
\label{q:dir1}The same direction must also cover agents that communicate in natural language, because multi-agent evidence beyond benchmarks is thin in the coded corpus (Section~\ref{sec:approaches}). Three components are needed. First, \emph{compositional assume--guarantee contracts}, in which each agent's guarantee is an obligation on its outputs under assumptions about its peers' outputs, stated over message classes or intents rather than fixed schemas~\citep{dennis2020verifiable,verifyingMulti2025}. Second, \emph{formal interaction protocols}: permitted speech acts, delegation and handoff rules, and termination and deadlock-freedom conditions that can be monitored at run time~\citep{formalizingDistributed2025}. Third, \emph{composition rules} stating when local guarantees jointly imply a system-level property and when they do not, for example under shared-resource contention or unsynchronized escalation. Candidate measures are contract-violation detection recall under inter-agent interference and composition-check coverage.
\endgroup

\textbf{2) Adversarial trajectory generation.} \emph{Which techniques can generate test trajectories that cover the fat-tail failure modes unique to agentic systems?} This direction \add{starts from} the \emph{Temporal $\times$ Benchmarks} cell in Table~\ref{tab:gap-matrix}, which contains only 4 coded appearances\add{: few papers whose primary validation object is temporal contribute a benchmark, although long-horizon benchmarks coded under other primary dimensions are not counted in this cell}. The \add{need is not more environments per se, but} generators for rare, off-policy, state-dependent trajectories in which memory errors, tool misuse, delayed side effects, and adversarial perturbations combine over time.

Existing environments provide useful starting points, including WebArena, WorkArena, and AgentDojo~\citep{webarena2024,agentdojo2024}, together with the adversarial Nemesis benchmark~\citep{nemesisAdversarial2025}, but they do not yet provide systematic coverage of temporally extended failure surfaces.\add{ Multi-agent settings additionally need generators that mutate inter-agent message order, timing, and role assignment, not only single-agent inputs.} \add{Physician red-teaming of patient-facing chatbots shows why: different models fail on largely different questions, and a recent health benchmark penalizes only the unsafe answers that clinicians recorded in advance~\citep{draelos2026}. What is needed is search procedures that preferentially discover tail failures through environment mutation, counterfactual replay, disturbance injection, and policy-guided stress generation.}

The expected artifact is a \emph{fat-tail generator}: a replayable test generator that emits adversarial trajectory families and minimal counterexamples for debugging and audit. Progress is visible through rare-event coverage, adversarial failure discovery, and replay reproducibility.

\textbf{3) Temporal runtime monitoring.} \emph{Which runtime monitors can detect trajectory-level drift early enough to trigger containment before localized deviations compound into unsafe outcomes?} This direction \add{starts from} the \emph{Temporal $\times$ CPS Validation} cell in Table~\ref{tab:gap-matrix}, which contains only \add{1 coded appearance}, \add{one of the sparsest cells of the matrix}. That scarcity matters because many agentic deployments are not one-shot services but closed-loop systems in which delayed tool effects, memory corruption, schema changes, and workload shift accumulate across operational time.

Three monitoring channels recur across the surveyed literature: memory and retrieved-context drift, tool-version and schema change, and environment or workload shift. The unsolved problem is to turn those signals into revalidation triggers, degradation modes, rollback rules, and escalation paths that activate before error propagation becomes opaque. Assurance monitoring and runtime verification provide the initial containment mechanisms~\citep{hawkins2021amlas,rtbas2025}; MLOps and expert-monitoring work identifies observable signals~\citep{mlops2023,expertmonitoring2024}. FDA \add{change-control }and MDCG \add{post-market surveillance }guidance \add{set the associated expectations for medical devices}~\citep{fda-pccp2025,mdcg2025-10}, while agent harnesses and evolving protocols make the need concrete~\citep{ahe2026,autogenesis2026}. A key implication of the Safety split in Section~\ref{sec:approaches} is that runtime monitors must distinguish security-triggered containment from functional-safety degradation.\add{ Whether prompt-injection defenses and graceful-degradation mechanisms compose, rather than interfere with each other, remains an open question.}

The expected artifact is an \emph{event calculus for revalidation}: a monitor specification that binds observable drift events to containment actions and evidence refresh. It specifies which changes in memory state, tool schema, workload, or policy behavior are revalidation-relevant, how those signals combine across time, and which assurance action each pattern triggers.\add{ For multi-agent deployments the event vocabulary must include interaction events (message loss, conflicting instructions, handoff timeouts) so that drift in joint behavior, not only per-agent drift, can trigger revalidation.} Progress can be assessed through drift-slope sensitivity, unsafe-action rate (UAR), and response-rule correctness: block, rollback, degrade, or escalate before downstream hazard propagation becomes opaque.

\textbf{4) Human oversight and evidence legibility.} \emph{Which oversight and evidence-chain designs allow human reviewers to verify, contest, and renew trajectory-level assurance claims as systems evolve?} This direction \add{starts from} the \emph{Regulatory $\times$ Runtime Monitors} cell in Table~\ref{tab:gap-matrix}, which contains only \add{6} coded appearances\add{: few regulatory-primary papers contribute a runtime monitor, and how monitored traces and documented regulatory claims can be connected in a reviewable chain remains an open design question}.

Two distinct concerns have to be separated here because they draw on different literatures and produce different artifacts. The first is \emph{human oversight}: human-computer interaction (HCI) and human-factors work asks when to interrupt, how to present uncertainty, what escalation path is usable, and how to avoid over-trust, alert fatigue, and nominal human-in-the-loop theater. Trust and human-factors research establishes the basic risks~\citep{lee2004trust,parasuraman1997humans}; \add{a healthcare-agent review and a physician red-teaming study of chatbot answers to patient questions indicate why they matter for clinical agents, although neither studies deployed agentic trajectories}~\citep{zhao2026,draelos2026}, and cascade analysis exposes their systemic consequences~\citep{sambasivan2021cascades}. The second is \emph{evidence legibility}: assurance-case, formal-methods, and traceability work asks whether a reviewer can reconstruct what claim was made, which evidence supported it, which monitor fired, which override occurred, and whether that evidence remained valid at decision time. FDA documents establish the lifecycle-evidence expectation~\citep{fda-action-plan,fda-gmlp}; IEC and IEEE guidance specify complementary traceability practices~\citep{iec62304,ieeeguide2025}. \add{The EU MDR provides the binding regulatory setting~\citep{eu-mdr}, which non-binding MDCG guidance explains~\citep{mdcg2025-6,mdcg2025-10}}, with assurance-case work addressing reviewable arguments~\citep{alelyani2025}. Treating them as one problem obscures the fact that a usable override interface is not yet an audit-ready assurance package, and conversely that a formally traceable log may still be unusable for real-time human review.

The expected artifacts are therefore an \emph{oversight interface specification} for escalation, override, acknowledgment, and handoff, plus \emph{proof objects} and linking schemas that bind specifications, test traces, runtime logs, overrides, and update events into audit-ready claims. Oversight quality is evaluated through reviewer uptake, correct override or escalation use, and low interruption burden; evidence legibility is evaluated through claim support, audit replay success, and freshness checks such as EFI.\add{ In multi-agent settings the evidence chain must also attribute each action to an agent and a delegation path, since causal attribution otherwise fails (Section~\ref{sec:taxonomy}).} Together, these directions shift validation from measuring whether a component can succeed to demonstrating that trajectories remain acceptable, governable, and evidentially current in context.

\begingroup
Two requirements cut across the four directions. First, validation for agentic systems bundles activities that are related but functionally different, and conflating them hides gaps (Table~\ref{tab:functions}). Observing a trajectory is not judging it, detecting an anomaly is not verifying a requirement, intervening is not recovering, and none of these is by itself evidence for an assurance claim. Second, the four directions need a traceable chain from requirements to claims (Table~\ref{tab:trace-chain}), so that each claim can be audited back to the hazard that motivated it and invalidated when a link changes.

\begin{table*}[!tbp]
\centering
\caption{Functionally distinct activities in agentic-system validation.}
\label{tab:functions}
\footnotesize
\setlength{\tabcolsep}{3pt}
\renewcommand{\arraystretch}{1.1}
\begin{tabularx}{\textwidth}{@{}>{\raggedright\arraybackslash}p{2.6cm}>{\raggedright\arraybackslash}p{3.4cm}>{\raggedright\arraybackslash}X>{\raggedright\arraybackslash}p{2.0cm}@{}}
\toprule
\textbf{Activity} & \textbf{Question answered} & \textbf{Typical input $\rightarrow$ output} & \textbf{Timing} \\
\midrule
Runtime observation & What is happening? & Execution events $\rightarrow$ structured trajectory record; no judgment made & Continuous \\
\midrule
Anomaly detection & Does this deviate from expected or learned behavior? & Trajectory record $\rightarrow$ score or alert; a statistical deviation, not a violated requirement & Online \\
\midrule
Formal verification & Does the model or specification satisfy a stated property for all modeled behaviors? & Model, property $\rightarrow$ proof or counterexample; limited to formalizable components & Design time; monitor synthesis \\
\midrule
Intervention & Which action limits harm now? & Monitor signal $\rightarrow$ block, flag, or escalate; needs authority and a latency budget & Online \\
\midrule
Recovery & How is a safe and valid state restored? & Post-intervention state $\rightarrow$ rollback, degraded mode, handoff, or revalidation & Online and offline \\
\midrule
Assurance-evidence generation & What evidence supports the claim, and is it still valid? & Outputs of the above $\rightarrow$ traceable evidence bundle tied to claims & Lifecycle \\
\bottomrule
\end{tabularx}
\end{table*}

\begin{table*}[!tbp]
\centering
\caption{Traceability chain from requirements to assurance claims, illustrated with the constructed medical trace (Table~\ref{tab:trace-medical}).}
\label{tab:trace-chain}
\footnotesize
\setlength{\tabcolsep}{3pt}
\renewcommand{\arraystretch}{1.1}
\begin{tabularx}{\textwidth}{@{}>{\raggedright\arraybackslash}p{3.0cm}>{\raggedright\arraybackslash}X>{\raggedright\arraybackslash}p{4.2cm}@{}}
\toprule
\textbf{Stage} & \textbf{Artifact and example} & \textbf{Links to} \\
\midrule
Hazard and requirement & H1: advice issued on outdated clinical data; R1: medication-relevant advice uses data no older than 60 minutes & Hazard analysis; regulatory obligation \\
\midrule
Test scenario & Stale-panel scenario; paraphrase-replay set for the triage classification & Covers R1 and the consistency requirement \\
\midrule
Observed trajectory & Trace $(s_t,a_t,m_t)$ with source timestamps in $m_t$ & Produced by scenarios and by field operation \\
\midrule
Runtime monitor & EFI check against the 60-minute limit; escalation-appropriateness rule & Monitors R1 on observed trajectories \\
\midrule
Intervention decision & Block pending refresh; escalate to clinical staff; decision logged & Triggered by monitor signals \\
\midrule
Assurance claim & ``Medication advice is issued only on evidence within 60 minutes,'' supported by scenario results, monitor logs, and intervention records; invalidated when the tool schema or freshness policy changes & Claims tie back to H1 and R1 and expire on change events \\
\bottomrule
\end{tabularx}
\end{table*}
\endgroup

\section{Open Challenges and Methodological Directions}
\label{sec:challenges}

Even with the proposed validation stack, several structural problems are likely to persist beyond any single research direction.

\textbf{Scalability of assurance.} Assurance cost \add{is likely to grow rapidly} as deployments expand from single agents to interacting ecologies of tools, humans, and organizations. Compositional methods can validate local properties, but system-level properties of multi-agent systems with dynamic membership and open communication vocabularies resist tractable state-space analysis~\citep{scienceScaling2025, harnessingMultiagent2025}. Better local validation does not automatically compose into system-level assurance, and each additional coordination or security layer adds obligations that do not aggregate cleanly~\citep{hybridMultiagent2025,decisionAlignment2025}. A general solution likely requires new assume-guarantee frameworks adapted to open-ended LLM interaction rather than bounded agent programs.

\textbf{Performance versus assurance overhead.} Richer simulation, denser logging, monitoring, and more frequent human review improve evidence depth, but each creates a persistent tension with latency, usability, compute efficiency, and workflow acceptance. Work on sustainable monitoring explicitly quantifies this trade-off between accuracy and energy efficiency~\citep{sustainablyMonitor2025}, while cost analyses of agentic reasoning show that assurance instrumentation carries measurable operational overhead~\citep{costDynamic2025}. Adaptive protocols that reduce monitoring granularity during low-risk operation
and scale up when risk indicators rise are a promising partial mitigation~\citep{adaptiveCommunication2025}, but the fundamental tension between evidentiary depth and performance remains unresolved.

\textbf{Human-AI teaming as a moving target.} Human-in-the-loop validation is a pillar of the research agenda, but it does not fully solve the problem that human behavior changes in response to automation. Over-trust, under-reliance, intervention fatigue, complacency, and shifting norms of responsibility are adaptive properties of socio-technical systems, not static interface issues. Mission-oriented and user-facing agent studies illustrate this shift~\citep{agenticMission2025,agentUser2025}, including personalized-agent settings~\citep{agentPersonalized2025}. Validating the agent alone therefore undershoots the problem; validating the human-agent team is difficult precisely because the target of validation evolves through use. Studies designed to capture this dynamic, such as tracking override behavior longitudinally, varying automation reliability systematically, and measuring responsibility diffusion after incidents, \add{are rare among the studies cited here}.

\textbf{Evidence portability and regulatory harmonization.} Even well-formed assurance artifacts may not transfer across domains, jurisdictions, or procurement settings. Regulatory concepts such as acceptable risk, traceability, and post-deployment accountability are not interpreted identically across governance regimes, and benchmarks and architectures keep changing faster than evaluation norms can stabilize. Verifiable-semantics and accountability-oriented architectures provide candidate foundations~\citep{verifiableSemantics2025,accountabilitybasedArchitectural2025}; agent fusion and team-aware methods extend them to multi-agent systems~\citep{agentfusionMultiagent2025,taaeTeamaware2025}, alongside explicit verification models~\citep{verificationModel2025}. The engineering challenge is therefore not only how to produce evidence, but how to make assurance claims portable, comparable, and durable under heterogeneous external oversight. This ultimately requires coordination between technical standards bodies and governance institutions that lies beyond any single engineering contribution.

The four-direction validation stack addresses engineering within a deployment lifecycle. Assurance at scale, overhead management, human-team dynamics, and evidence portability remain broader socio-technical and governance challenges.

\section{Conclusion}
\label{sec:conclusion}

This survey argues that unit, integration, and benchmark results alone do not provide adequate validation coverage for agentic systems. Their relevant behavior is a trajectory realized through planning, tool use, memory, coordination, and adaptation in context. That shift changes the assurance target from isolated component correctness to bounded acceptable behavior over time.

Drawing on a corpus of \add{262} papers, the survey makes four claims. First, classical testing abstractions remain necessary but stop short of the validation target once software acts autonomously in consequential environments. Second, the literature is most usefully organized by five validation dimensions: behavioral, safety, temporal, regulatory, and multi-agent, each capturing distinct failure modes and evidence needs. \add{Third, the reviewed literature is unevenly distributed across those dimensions: behavioral evaluation is the largest primary category (105 of 262 papers, 97 of which carry the Benchmarks tag), temporal validity is the least represented (14 papers), regulatory evidence consists mainly of assurance cases and regulatory analysis and is concentrated in IEEE-retrieved, peer-reviewed work, and multi-agent evidence beyond benchmarks is thin.} Fourth, the survey translates those asymmetries into a four-direction lifecycle stack spanning bounded-autonomy specifications\add{ (including compositional contracts for multi-agent systems)}, adversarial trajectory generation, temporal runtime monitoring, and human-verified evidence chains, an interpretive synthesis of the surveyed gaps motivated by the corpus-level counts.

The broader implication is methodological as much as technical. Validation for agentic AI is no longer mainly about measuring whether a model can succeed on a task; it is about demonstrating that trajectories remain acceptable, inspectable, and governable as environments, interfaces, and evidence change. That is why agentic AI demands not a larger benchmark alone, but a new validation paradigm.

\section*{Statements and Declarations}

\subsection*{Funding}
This research received no external funding.

\subsection*{Ethics approval and consent to participate }
Not applicable

\subsection*{Consent for publication}
Not applicable

\subsection*{Competing interests}
The authors declare that they have no conflict of interest related to this work.

\subsection*{Data availability}
\add{The search queries, title lists of the date-filtered retrieval exports and the retrieval log, included-study and candidate eligibility records (555 candidates), codebook, coding sheets, bibliography matching record, case-study claim records, analysis outputs, and the script that recomputes the derived tables are available at \repourl.}

\subsection*{Author contributions}
F.O.M. conceived the original idea, designed the review methodology, developed the five-dimensional taxonomy, conducted the literature review and analysis, synthesized the findings, prepared the case studies, and drafted the manuscript. L.D. and G.T. contributed to the methodology, literature analysis, taxonomy refinement, interpretation of the findings, and critical revision of the manuscript. S.S., F.L., A.P., and G.M. supervised the work, contributed to the interpretation of the findings, and critically revised the manuscript for important intellectual content. All authors read and approved the final manuscript.

\expandafter\def\csname b@label\endcsname{}\nocite{label}
\bibliography{sn-bibliography}

\end{document}